\documentclass[10pt,twocolumn,letterpaper]{article}

\usepackage[pagenumbers]{iccv} 

\usepackage{subcaption}
\usepackage{etoolbox, array}
\usepackage[font=small,labelfont=bf,aboveskip=3pt,belowskip=-5pt]{caption}

\newdimen\abovecrulesep
\newdimen\belowcrulesep
\makeatletter
\patchcmd{\@@@cmidrule}{\aboverulesep}{\abovecrulesep}{}{}
\patchcmd{\@xcmidrule}{\belowrulesep}{\belowcrulesep}{}{}

\definecolor{demphcolor}{RGB}{144, 144, 144}
\definecolor{mygray}{gray}{0.4}
\definecolor{lightgray}{rgb}{0.9, 0.9, 0.9}

\newlength\savewidth

\newcommand{\tablestyle}[2]{\setlength{\tabcolsep}{#1}\renewcommand{\arraystretch}{#2}\centering\footnotesize}
\makeatletter\renewcommand\paragraph{\@startsection{paragraph}{4}{\z@}{.5em\@plus1ex\@minus.2ex}{-.5em}{\normalfont\normalsize\bfseries}}
\makeatother

\definecolor{cvprblue}{rgb}{0.21,0.49,0.74}
\usepackage[pagebackref,breaklinks,colorlinks,allcolors=cvprblue]{hyperref}
\usepackage{multirow}
\usepackage{graphicx}
\usepackage{subcaption}
\usepackage{algorithm}
\usepackage{tcolorbox}
\usepackage{bm}
\usepackage{wrapfig}
\usepackage{booktabs}
\usepackage[noend]{algorithmic}

\usepackage[normalem]{ulem}

\newcommand{\ours}{VisVM\xspace}

\def\paperID{6589} 
\def\confName{ICCV}
\def\confYear{2025}

\title{Scaling Inference-Time Search with Vision Value Model \\ for Improved Visual Comprehension}

\author{%
  Xiyao Wang$^{1, 2, \dag}$, Zhengyuan Yang$^{2}$, Linjie Li$^{2}$, Hongjin Lu$^{1}$, Yuancheng Xu$^{1}$ \\
  Chung-Ching Lin$^{2}$, Kevin Lin$^{2}$, Furong Huang$^{1, \ddag}$, Lijuan Wang$^{2, \ddag}$ \\
  $^1$University of Maryland, College Park \quad $^2$Microsoft \\
  $^\dag$\texttt{xywang@umd.edu} \qquad $^\ddag$\texttt{Equal advise}\\
}
\usepackage{tikz}
\newcommand*\circled[1]{\tikz[baseline=(char.base)]{
            \node[shape=circle,draw,inner sep=1.5pt] (char) {#1};}}
            
\begin{document}

\twocolumn[{\renewcommand\twocolumn[1][]{#1}
\maketitle
\centering
\centerline{\includegraphics[width=1.04\linewidth]{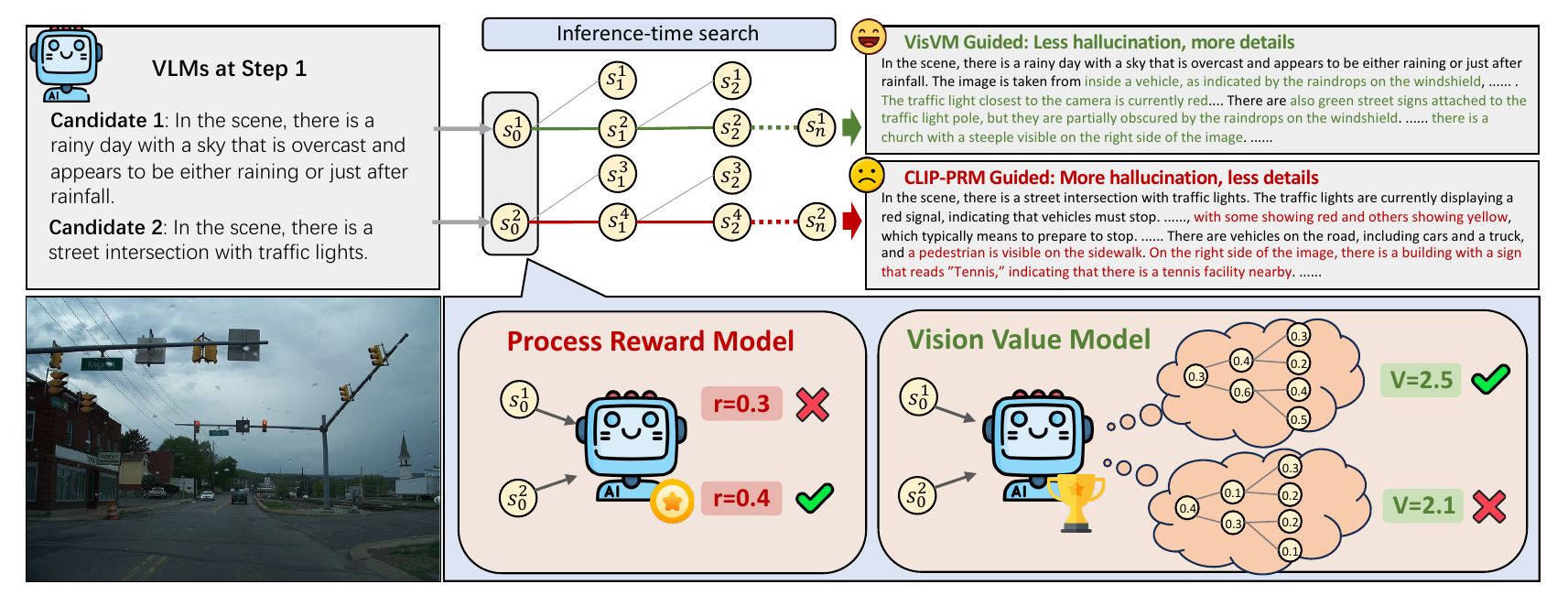}}
\captionof{figure}{An illustration of how \ours can better guide vision language model (VLM) during inference-time search. When selecting response candidates at each step, the process reward model (PRM) only considers the immediate reward, whereas \ours predicts the long-term value by considering potential hallucinations in subsequent generated sentences. This enables \ours to avoid response candidates with higher hallucination risks and generate image descriptions that are less prone to hallucination and more detailed.}
\label{fig:teaser}
\vspace{3em}
}]

\begin{abstract}
\vspace{-16pt}




Despite significant advancements in vision-language models (VLMs), there lack effective approaches to enhance response quality by scaling inference-time computation. This capability is known to be a core step towards the self-improving models in recent large language model studies. 
In this paper, we present \textbf{Vis}ion \textbf{V}alue \textbf{M}odel (\ours) that can guide VLM inference-time search to generate responses with better visual comprehension.
Specifically, \ours not only evaluates the generated sentence quality in the current search step, but also anticipates the quality of subsequent sentences that may result from the current step, thus providing a long-term value. In this way, \ours steers VLMs away from generating sentences prone to hallucinations or insufficient detail, thereby producing higher quality responses.
Experimental results demonstrate that \ours-guided search significantly enhances VLMs' ability to generate descriptive captions with richer visual details and fewer hallucinations, compared with greedy decoding and search methods with other visual reward signals. 
Furthermore, we find that self-training the model with the \ours-guided captions improves VLM's performance across a wide range of multimodal benchmarks, indicating the potential for developing self-improving VLMs.


\end{abstract}    
\section{Introduction}
\label{sec:intro}

Vision language models (VLMs) have advanced rapidly, excelling in multimodal tasks involving single images~\cite{llava, bai2023qwen, chen2024internvl, shi2024eagle}, multiple images~\cite{jiang2024mantisinterleavedmultiimageinstruction, li2024llavanextinterleavetacklingmultiimagevideo}, and videos~\cite{li2024llavaonevisioneasyvisualtask, xue2024longvilascalinglongcontextvisual, wang2024qwen2vlenhancingvisionlanguagemodels}. These capabilities stem from large-scale, high-quality training data, often sourced from web-crawled image-text pairs~\cite{radford2021learning,jia2021scaling} with effective filtering~\cite{changpinyo2021conceptual,yuan2021florence,hu2021scaling}, or enriched through techniques like distillation from stronger VLMs~\cite{chen2023sharegpt4v}, human annotations~\cite{betker2023improving}, or added textual descriptions~\cite{lai2025veclip}. Despite this progress, VLMs still suffer from visual hallucinations~\cite{liu2023aligning,guan2024hallusionbenchadvanceddiagnosticsuite,wang-etal-2024-mementos} and often neglect less salient image regions, limiting their real-world utility. While increasing the scale and quality of training data could help, this approach incurs significant annotation and API costs, making it less scalable. This raises a key question: \emph{Can we enhance VLMs' response quality at inference time, and leverage these improved responses to further advance VLMs' visual comprehension?} 

Recent studies on large language models (LLMs)~\cite{o1blog,lightman2023letsverifystepstep, snell2024scalingllmtesttimecompute, yang2024qwen25mathtechnicalreportmathematical,snell2024scaling} highlight inference-time search as a promising approach for improving response quality, complementary to training time effort. By leveraging a pretrained process reward model~\cite{zhang2024rest, tian2024toward}, LLMs can perform search iterations to produce high-quality outputs, with these refined responses showing potential as synthetic training data to enhance reasoning capabilities. However, extending this approach to VLMs for improved visual comprehension poses unique challenges, particularly in defining a reward signal. While process and outcome rewards are relatively straightforward for LLM tasks like coding and math, VLM tasks—such as descriptive captioning—lack clear outcome measures and require cohesive paragraph image descriptions that consist of multiple global and regional caption sentences. In these cases, each sentence must not only be accurate locally but also contribute to a coherent overall response.

To this end, we propose the \textbf{Vis}ion \textbf{V}alue \textbf{M}odel (\ours), a value network to guide VLM inference-time search by generating descriptive captions in a step-by-step manner, with each step producing one sentence. 
As shown in Figure~\ref{fig:teaser}, \ours takes the image and generated sentence at each step as inputs, predicting a \textbf{long-term value} to ensure both visual-text alignment and coherence. 
\ours is grounded in two key insights that distinguish it from traditional process reward models in LLM literature~\cite{uesato2022solving,lightman2023let,cobbe2021training,hosseini2024vstartrainingverifiersselftaught,wang2024mathshepherdverifyreinforcellms}: \textbf{(1) Forward-looking coherence:} Unlike approaches that rely solely on the local reward of the current sentence, \ours predicts future consequences to maintain global consistency. It is trained using Temporal Difference (TD) learning~\cite{sutton1988learning}, enabling it to assess long-term effects rather than just evaluating immediate responses. This forward-looking signal helps mitigate hallucinations by preventing sentences that may lead to inconsistencies in subsequent steps.
\textbf{(2) Comprehensive visual grounding:} To reduce hallucinations, the reward signal must encapsulate rich visual semantics. We achieve this by leveraging CLIP's text-image similarity metric, which effectively captures visual concepts and enforces alignment.

\begin{figure}[!tbp]
    \centering
    \begin{subfigure}{\linewidth}
        \includegraphics[width=1.02\linewidth]{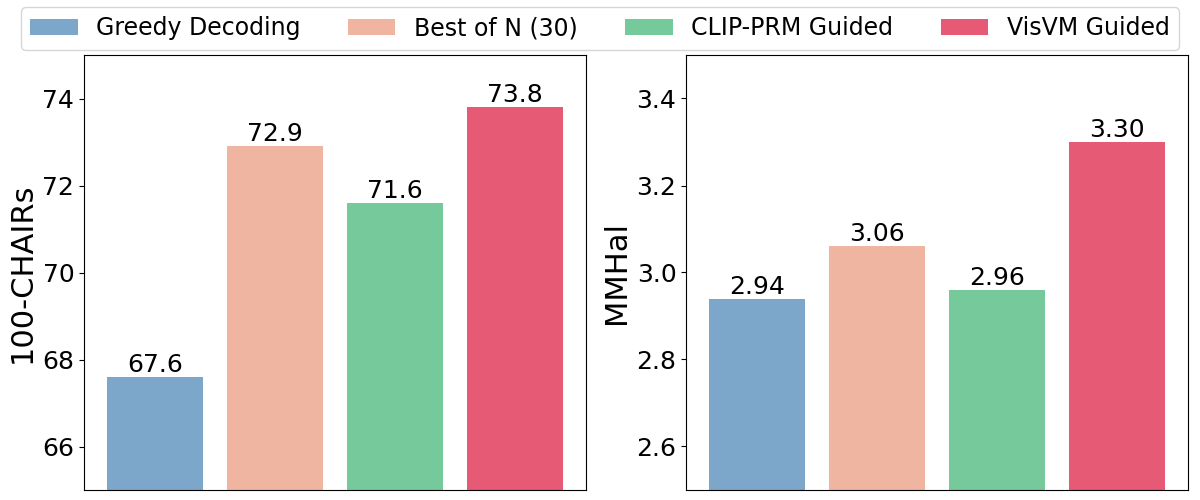}
        \label{fig:infer_bar}
    \end{subfigure}
    \begin{subfigure}{\linewidth}
        \includegraphics[width=\linewidth]{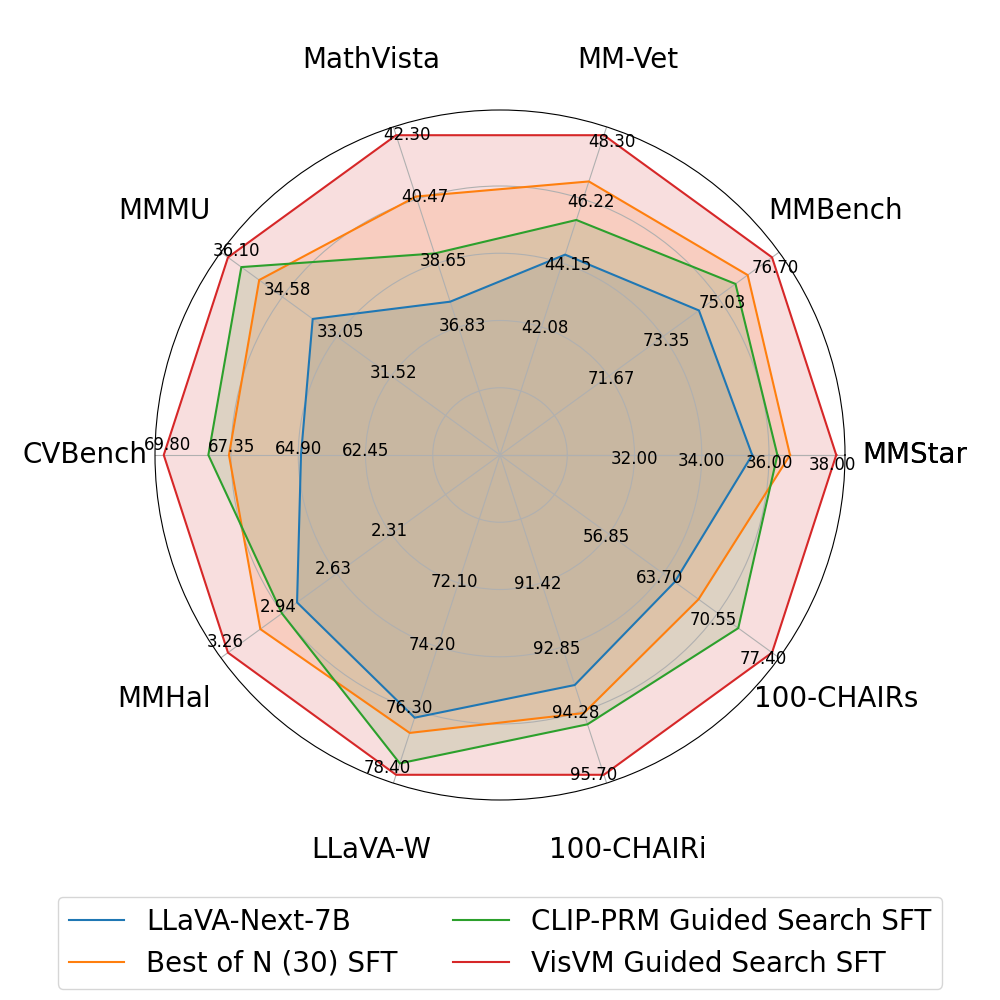}
        \label{fig:main_res}
    \end{subfigure}
    \caption{\textbf{Upper}: CHAIRs and MMHal score of descriptive captions generated by LLaVA-Next-7B during inference-time using different search methods. \ours-guided search clearly outperforms other methods, indicating reduced visual hallucinations. Notably, even with a smaller search budget (search size 6 vs. search size 30), our approach still surpasses the Best-of-N method. 
    \textbf{Lower}: Comparisons of LLaVA-Next-7B after fine-tuning with descriptive captions from different search methods, with \ours-guided search achieving favorable results across all 9 benchmarks.}
    \label{fig:combined}
    \vspace{-10pt}
\end{figure}

We validate the effectiveness of \ours through two main experiments: inference-time \ours-guided search and self-improvement training. 
\textbf{(1)} Using \ours as a guidance signal for VLM inference-time search to generate descriptive image captions, we observe a substantial reduction in hallucinations and more detailed image descriptions. 
In both GPT and human evaluations, captions generated with \ours consistently outperform those produced by greedy decoding, best-of-N decoding, and CLIP-PRM-guided search.
Notably, \ours-guided captions are preferred 74\% of the time over those from greedy decoding.
\textbf{(2)} To better leverage \ours's inference-time enhancement of VLM responses, we use \ours-guided captions as the Supervised Fine-Tuning (SFT) data to self-train the original VLM (LLaVA-Next-7B and Qwen2-VL-7B). Across nine standard benchmarks, \ours-guided self-training improves the performance of the original VLMs by an average of 10.8\% and 7.3\%, respectively.

Our contribution can be summarized as follows:
\begin{itemize}
\setlength\itemsep{3pt}
\item
We introduce \ours, a stepwise value model designed to provide long-term vision value signals to guide VLM inference-time search. 
To the best of our knowledge, \ours is the first exploration into enhancing VLM visual comprehension through inference-time search.
\item
\ours-guided search effectively reduces visual hallucinations and enriches image descriptions with more visual detail, by increasing the inference-time computation.
\item
Descriptive captions generated by \ours-guided search can be leveraged as high-quality SFT data, forming a robust self-training pipeline that significantly enhances VLM visual comprehension across 9 benchmarks.
\end{itemize}

\section{Related Work}
\vspace{1mm}
\noindent\textbf{Vision language models.} 
Significant advances~\cite{radford2021learning,yuan2021florence,wang2021simvlm,wang2022git,yu2022coca,li2024multimodal} have been made on vision-language modeling, which jointly understands the visual and text inputs for various tasks such as image captioning~\cite{chen2015microsoft} and visual question answering~\cite{vqav2}. Recently, modern vision language models~\cite{alayrac2022flamingo,gpt4v,yang2023dawn,llava,wang2023cogvlm,chen2024internvl,bai2023qwen,geminiteam2023gemini} further combine multimodal modeling with large language models to enable stronger capabilities, such as instruction following, in-context learning, and zero-shot generalization. 
However, VLMs still exhibit the issue of hallucination~\cite{guan2024hallusionbenchadvanceddiagnosticsuite,wang-etal-2024-mementos,wang2023evaluation}. 
Existing work mitigates hallucination in VLMs by improving the quality of SFT data~\cite{wang2023mitigatingfinegrainedhallucinationfinetuning,chen2023mitigatinghallucinationvisuallanguage} or through post-training methods~\cite{zhou2024aligning,liu2023aligning,wang2024enhancing,sun2023aligninglargemultimodalmodels}. In this paper, we explore reducing hallucination in responses not through training but by using inference-time search to improve the quality of responses.

\vspace{1mm}
\noindent\textbf{Descriptive captioning.} 
Descriptive captioning aims to describe each image with a long, comprehensive text paragraph. Recent studies show the effectiveness of using synthetic descriptive captions for vision language model. The pairs of images and paragraph captions can be used for image-to-text understanding models~\cite{chen2023sharegpt4v,wang2023cogvlm}, text-to-image generation models~\cite{betker2023improving,esser2024scaling}, as well as image-text contrastive models~\cite{lai2025veclip,wu2024lotlip,lai2024revisit}.
In this study, we focus on improving the descriptive caption quality of a trained VLM by exploring effective approaches to scale the inference-time search.

\vspace{1mm}
\noindent\textbf{Inference-time search.} 
Inference-time search strategies have proven crucial for complex reasoning and planning tasks in robotics~\cite{wang2023coplanner, hansen2022temporaldifferencelearningmodel}, chess~\cite{silver2016mastering}, and autonomous driving~\cite{Teng_2023}. 
The advent of OpenAI-O1 has further advanced inference-time search within LLMs. 
By applying various search techniques in the language space, such as controlled decoding~\cite{chakraborty2024transfer, xu2024genarm}, best of N~\cite{lightman2023letsverifystepstep, li2024common7blanguagemodels}, and Monte Carlo tree search~\cite{zhang2024rest, tian2024toward, wang2024litesearch, wang2024towards}, LLMs achieve better model responses, thus enhancing performance. 
A good process reward model (PRM) is essential during inference-time search, as the quality of the reward signal determines the quality of the responses found and the budget required to achieve high-quality responses. 
Various PRMs~\cite{uesato2022solving,lightman2023let,cobbe2021training,hosseini2024vstartrainingverifiersselftaught,wang2024mathshepherdverifyreinforcellms} have been proposed in LLMs to address mathematical and coding problems. 
Moreover, \citet{brown2024large} and \citet{snell2024scaling} have found that scaling the search budget during inference time can further enhance LLM performance. 
However, inference-time search remains underexplored in VLMs. 
\citet{zhou2024calibrated} proposed using CLIP as a signal for generating positive and negative samples post-training, but did not further investigate its impact as a PRM on VLM inference-time search. 
In this paper, we propose a vision value model superior to CLIP as a search signal for inference-time search, aimed at enhancing the visual comprehension abilities of VLMs.
\section{Vision Value Model}

In this section, we introduce the proposed Visual Value Model (\ours). We first present the problem formulation of large multimodal model (VLM) inference in Section~\ref{sec 3.1}, and then discuss the training process for \ours  in Section~\ref{sec 3.2}. Section~\ref{sec 3.3} shows how to employ \ours for effective inference-time search in VLMs. 
 
\subsection{Formulation of VLM Inference}
\label{sec 3.1}
We first introduce the formulation of VLM inference. 
We consider an VLM characterized by probability distribution $p_\theta$, represented as the policy $\pi_\theta$. This model processes a prompt-image pair ($x, I$) as input to generate a response $\bm{y} = [y_1, y_2,... , y_m]$, where $y$ consists of $m$ step-level responses.
Each step-level response $y_i$ is treated as a sample drawn from the conditional probability distribution 
$y_i = p_\theta(\cdot|x, I, \bm{y}_{<i})$. 
In this paper, we define each step-level response as sentence-level, meaning that at each step, the output is a single sentence.
Consequently, the text generation task can be formulated as an Markov Decision Process (MDP) problem defined by the tuple $(\mathcal{S}, \mathcal{A}, \mathcal{R}, \gamma)$.
$\mathcal{S}$ is the state space. Each state is defined as a combination of the generated sentences and the image. The initial state $s_0$ corresponds to  image $I$ and input prompt $x$. 
$\mathcal{A}$ is the action space where each action is the sentence generated in that step. 
We also have the reward function $\mathcal{R}$ to evaluate the reward of each action, which is also known as process reward model (PRM) in LLMs. 
$\gamma$ denotes the discount factor.
With this MDP modeling, we can search additional states by increasing the inference-time compute, thereby obtaining a better VLM response $y$. The core of our method lies in the exploration of a better value model, namely \ours, which can better guide the inference-time search.

\subsection{\ours Training}
\label{sec 3.2}

\paragraph{Training method.} 
The primary goal of \ours is to estimate the long-term value of the current image-conditioned sentence in potential future sentence generation scenarios. 
To achieve this, we employ Temporal Difference (TD) learning~\cite{sutton1988learning}, a popular method in reinforcement learning, to train \ours for predicting the long-term vision value $V_\rho(y_i,I)$ at each state $s_i=(y_i,I)$. 
For a given triplet consisting of the current sentence $y_i$, the next sentence $y_{i+1}$, and an associated image $I$, we first use the PRM to estimate the reward $r_{s_i}$ of the current state $s_i$. We then train \ours using the following loss function, ensuring the predicted value for the current state $s_i$ matches the sum of the actual received reward and the discounted predicted value for the next state:
\begin{equation}
\label{global_density}
\resizebox{0.9\columnwidth}{!}{
$L(\rho) = -\mathbb{E}_{(y_i, y_{i+1}, I)\sim\mathcal{D}}\left( r_{s_i} + \gamma V_\rho(y_{i+1}, I) - V_\rho(y_i, I) \right)^2,$
}
\end{equation}
where $\gamma$ denotes the discount factor, $\rho$ is the learnable parameters of \ours, and $\mathcal{D}$ is our constructed training data.

\paragraph{Training data.} 
Training \ours requires the triplet of the current sentence, the next sentence, and an associated image. Such triplets can be extracted from pairs of images $I$ and paragraph descriptions $\bm{y} = [y_1, y_2,... , y_m]$.
It is imperative to generate a diverse set of responses using VLMs to explore potential subsequent sentences that each initial sentence may encounter, thereby accurately modeling the sentence's long-term value. 
We sample 9,215 images from the COCO 2017 training dataset and utilize the nine prompts from the LLaVA-150K dataset designed for description captioning. These prompts are randomly paired with the images to construct prompt-image pairs.
For each prompt-image pair, we generate five distinct responses using the VLM, using both greedy decoding and temperature decoding with temperature values set at different scales. 
After generating the paragraphs, each response is decomposed into sentence pairs consisting of the current sentence, the subsequent sentence, and the associated image.  
The final dataset $\mathcal{D}$, containing 378k samples, is used for training \ours.
We provide more training details in Appendix~\ref{app: C}.

\paragraph{Implementation details.} 
For the implementation of \ours, we take \textbf{LLaVA-Next-Mistral-7B} as our base model for an example. The implementation of \ours in the experiment section for both LLaVA-OV-7B and Qwen2-VL-7B follows the same procedures.
We concatenate a linear layer as the value head on top of the penultimate layer of LLaVA-Next-Mistral-7B. The output of this value head is a single scalar representing the cumulative reward, or long-term value, of all potential responses based on the current sentence and its paired image. Additionally, we initialize all parameters of \ours, except for this value head, using the parameters of LLaVA-Next-Mistral-7B.
For the training data, we use the base model corresponding to \ours to generate descriptions for all images and decompose them into training data.

For the PRM used in \ours training, we choose each VLM's vision encoder. For LLaVA-Next-Mistral-7B and Qwen2-VL-7B, we use CLIP-ViT, while for LLaVA-OV-7B, we use SigLIP. There are two main reasons for this:
\textbf{(1)} CLIP-like neural networks effectively measure the alignment between image content and text content by computing the similarity between image and text embeddings, making it highly suitable as PRM for visual comprehension task. Its effectiveness has also been demonstrated in prior studies~\cite{zhou2024calibrated}. 
\textbf{(2)} Additionally, since CLIP-VIT and SigLIP are the native visual encoders in base VLMs, using them as PRM eliminates the need for external models or human annotators. This self-rewarding mechanism is not only effective but also reduces costs.

\subsection{Inference-time Search using \ours}
\label{sec 3.3}

After training \ours, we use it as the signal to guide the VLM inference-time search for generating higher-quality responses. To encourage diversity among response candidates at each step of the search, we implement temperature decoding using $N$ distinct temperature configurations $T_n$. Given the current VLM as the policy $\pi_\theta$, it generates a conditional probability distribution $p_\theta(\cdot|x, I, \bm{y}_{<i}, T_n)$ based on the input image, prompt, temperature configuration, and previous step responses. We then sample $K$ responses from each $p_\theta$, yielding $N\times K$ response candidates for the current step. Each candidate's value is estimated using \ours, and the candidate with the highest value is selected as the response for the current step. This process continues iteratively until the complete response sequence is generated, \ie, only the EOS token is generated for the next sentence. The pseudo code for this search process is in Algorithm~\ref{alg: VisVM}.

\begin{algorithm}
\caption{\ours-Guided Inference-time Search}
\label{alg: VisVM}
\begin{algorithmic}[1]
\REQUIRE{
Test sample $\{ x, I \}$, 
VLM $p_\theta$, VisVM $V_\rho$, Step size $K$, Temperature configuration list $T$, Response $\bm{y}=[\ ]$}
\WHILE{Generation is not Done}
\STATE Current step response $y_i = \text{None}$, Current step max value $V_i^{max}=-$$\infty$
\FOR{temperature $T_n$ in $T$}
    \FOR{$k = 1, \ldots, K$}
    \STATE Generate response of the new step $j$: \\ $y_i^j=p_\theta(\cdot|x, I, \bm{y}_{<i}, T_n)$,
    \STATE Estimate step value $V_i^j=V_\rho(y_i^j, I)$,
        \IF{$V_i^j>V_i^{max}$}
        \STATE Current step max value $V_i^{max}=V_i^j$, 
        \STATE Current step response $y_i=y_i^j$
        \ENDIF
    \ENDFOR
\ENDFOR
\STATE Append current step response $y_i$ to $\bm{y}$
\ENDWHILE
\RETURN{Final response $\bm{y}$}
\end{algorithmic}
\end{algorithm}

\begin{figure*}[!htbp]
    \centering
    \begin{subfigure}[b]{0.5\linewidth}
        \includegraphics[width=\linewidth]{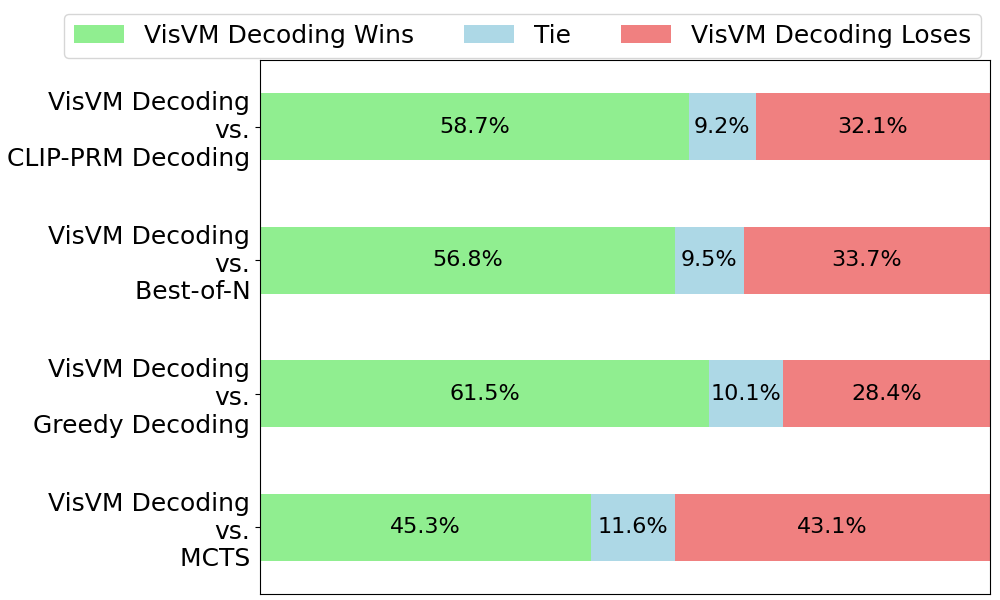}
        \caption{Win rate of \ours-guided search compared with other methods}
        \label{fig:gpt_eval}
    \end{subfigure}
    \hfill
    \begin{subfigure}[b]{0.45\linewidth}
        \includegraphics[width=\linewidth]{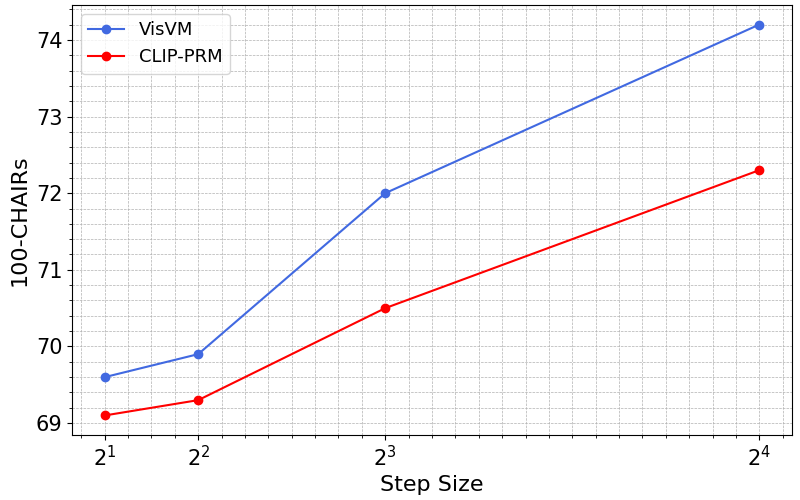}
        \caption{Scaling curve of search step size.}
        \label{fig:scaling_curve}
    \end{subfigure}
    \vspace{5pt}
    \caption{\textbf{(a)} Win rate of image descriptions generated using LLaVA-Next-7B with \ours-guided search compared with other search methods. We use GPT-4o api for evaluation. We can find \ours-guided search generated description significantly better than others methods. \textbf{(b)} Step size scaling curve for \ours-guided search and CLIP-PRM guided search. We report the CHAIRs score of image descriptions under different step sizes. \ours-guided search is 2$\times$ efficient than CLIP-PRM guided search.}
    \vspace{-10pt}
    \label{fig:combined_exp}
\end{figure*}

\section{Experiment}

In this section, we conduct experiments to answer the following two questions: 
1. Does the \ours-guided search yield higher-quality responses compared with other inference-time search methods (Section~\ref{sec_41})? 
2. Can the \ours-guided search be leveraged to generate high-quality SFT data, thereby improving the visual comprehension capabilities of VLMs through self-training (Section~\ref{sec_42})?

\subsection{Inference-Time Search with \ours}
\label{sec_41}
\subsubsection*{Baselines and Implementation Details}
In this section, we evaluate the ability of \ours on enhancing the response quality of VLMs by comparing its inference-time performance with various search methods. 
All experiments are based on \textbf{LLaVA-Next-Mistral-7B}. 
We consider the following baselines for inference-time search: 
(1) \textbf{Greedy decoding}: 
The standard decoding approach used for VLM decoding, where the responses with the highest probability are selected for each step. 
(2) \textbf{Best-of-N (BoN) decoding}: 
A widely used method to improve the quality of model responses during inference. For each prompt-image pair, we set five different temperature parameters [0.1, 0.3, 0.5, 0.7, 0.9] and generate six different model responses for each parameter, resulting in a total of 30 responses ($N=30$). We then use GPT-4o to select the best out of these 30 responses as the final response. 
(3) \textbf{CLIP-PRM guided search}: 
This method uses CLIP-ViT as the PRM to guide search. Since CLIP-ViT also serves as the reward model for training \ours, comparing \ours-guided search with CLIP-PRM guided search serves as the fair-comparison baseline.For CLIP-PRM guided search, we adopt the same search method as described in Section~\ref{sec 3.3}, with the only difference being that the guided signal is replaced by the CLIP similarity. All hyperparameters are kept identical to those used for \ours-guided search to ensure a fair comparison. 
We use temperature decoding with five different temperatures and greedy decoding to generate response candidates at each search step with a step size of 1, leading to six different response candidates per search step. The list of temperature configuration includes [0.1, 0.3, 0.5, 0.7, 0.9].
(4) \textbf{Monte Carlo Tree Search (MCTS)}: 
MCTS is a widely used inference-time search method for enhancing the performance of LLM. Thus, we adopt MCTS for VLM as another key baseline. To ensure a fair comparison, we continue to use CLIP-ViT as the PRM. At each step, during the expansion of child nodes, we generate six child nodes using five different temperature values along with greedy decoding. The number of MCTS iterations is set to 10.
We also provide comparison with other finetuning and decoding methods in Appendix~\ref{app: more_exp} due to space limitation.

\subsubsection*{\circled{1} \ours-Guided Search Improves Response Quality}
\label{sec: 4.1.1}
We sample 1,000 images from the COCO Train2017 dataset and randomly pair each image with 9 prompts from the LLaVA-150k detailed description dataset. This process results in 1,000 prompt-image pairs as an evaluation dataset. We use our method and three search baselines to generate a detailed descriptive caption for each pair.
Then, we pair captions generated by \ours-guided search with those from other decoding methods for the same image and subsequently assess the quality of the descriptions. 

\noindent \textbf{GPT evaluation.}
We use GPT-4o to compare \ours-guided search against other baselines,
as shown in Figure~\ref{fig:gpt_eval}. The prompt used for evaluation is in Appendix~\ref{app: A}. 
We observe a notable superiority in the win rate of the \ours-guided search compared with CLIP-PRM, BoN, and Greedy, with the win rate of 58.7\%, 56.8\%, and 61.5\%. 
Under GPT-based evaluation, while the advantage of \ours-guided search over MCTS is less pronounced than against other baselines, it still achieves a higher win rate, outperforming MCTS with 45.3\% compared to 43.1\%.

\noindent \textbf{Human evaluation.} 
We randomly select 200 prompt-image pairs and corresponding captions for human evaluation. We recruit 10 human evaluators to perform blind selections between these pairs to calculate the win rates of each method. We average their evaluations to obtain the final result in Table~\ref{tab: human eval}.
We find that descriptions generated by \ours-guided search are significantly preferred over those from CLIP-PRM, BoN, and Greedy decoding, with win rates of 62.4\%, 60.2\%, and 75.8\%, respectively. Compared to GPT-based evaluation, \ours exhibits a clearer advantage over MCTS under human evaluation. Human evaluators report more instances where captions from \ours and MCTS are of comparable quality, leading to a higher tie rate.
Notably, under the more reliable human evaluation, \ours achieves a 44.9\% win rate and a 15.2\% tie rate against MCTS, reinforcing its effectiveness.

Results from both GPT and human evaluations consistently demonstrate that \ours-guided search substantially enhances the response quality of VLMs in captioning.

\begin{figure}[!t]
    \centering
    \includegraphics[width=0.9\linewidth]{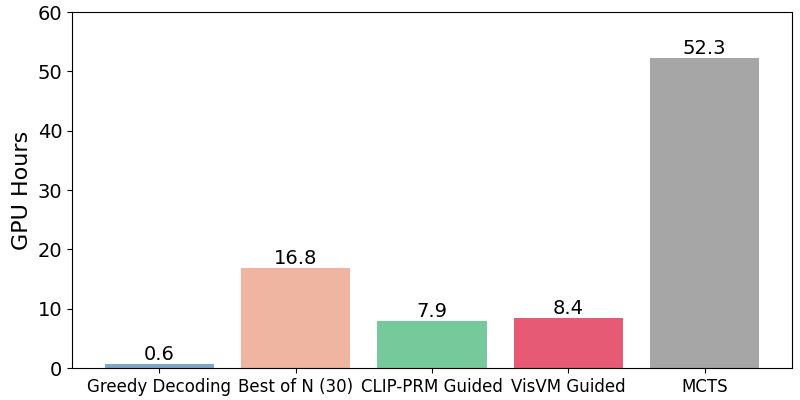} 
    \caption{Comparison of GPU hours required to generate 1000 image captions by different test-time compute methods. The GPU hour consumed by \ours-guided search is significantly lower than Best-of-N and MCTS.}
    \label{fig:gpu_hours}
\end{figure}

\noindent\textbf{Computational cost.}
We further compare the computational cost of various test-time compute methods to demonstrate the superiority of our approach. Specifically, we measure GPU hours required by each method to generate 1,000 image captions, utilizing an 8$\times$80GB A100 GPUs setup. The results, summarized in Figure~\ref{fig:gpu_hours}, reveal that all test-time compute methods significantly increased computational cost compared to greedy decoding. Among these, CLIP-guided search and \ours-guided search incur the most minor increases. 
Furthermore, under identical step size and temperature settings, while MCTS achieves performance comparable to \ours, it demands approximately seven times more computational resources. Additionally, MCTS must relearn the value function for each new prompt-image pair, highlighting its limited generalization capability. These findings further underscore the efficiency and effectiveness of \ours, demonstrating its superiority in both performance and scalability.

\begin{table}[!tb]
\centering
\caption{Human evaluation over 200 image-text pairs. \ours guided search still far surpasses other search methods, displaying results consistent with GPT evaluation.}
\label{tab: human eval}
\tablestyle{9pt}{1.1} 
\resizebox{\linewidth}{!}{%
\begin{tabular}{l|ccc}
\toprule 
Method & \ours wins & Tie & \ours loses\\
\midrule
vs. CLIP-PRM & 62.4\%    & 6.7\%   & 30.9\% \\
vs. MCTS     & 44.9\%    & 15.2\%  & 39.9\%\\
vs. BoN      & 60.2\%    & 9.6\%   & 30.2\% \\
vs. Greedy   & 75.8\%    & 5.4\%   & 18.8\% \\
\bottomrule
\end{tabular}
}
\vspace{-10pt}
\end{table}

\begin{table}[!htb]
\centering
\caption{Hallucination evaluation results using different inference-time searching on CHAIR and MMHal. \ours guided search achieves the best results, demonstrating strong capabilities in mitigating inference-time hallucination.}
\label{tab: infer_eval}
\tablestyle{8pt}{1.1} 
\resizebox{\linewidth}{!}{%
\begin{tabular}{l|l|cccccc}
\toprule 
Base & Searching Method & CHAIRs $\downarrow$ & CHAIRi $\downarrow$ & MMHal $\uparrow$ & MMHal rate $\downarrow$ & AMBER Cov $\uparrow$\\
\midrule
\multirow{5}{*}{LLaVA-Next-7B}  & Greedy (Default)    & 32.4 & 5.9 & 2.94 & 0.52 & 63.9\\
                                & MCTS                & 25.9 & 4.7 & 3.24 & 0.37 & 67.3\\
                                & BoN                 & 27.1 & 5.2 & 3.06 & 0.45 & 65.3\\
                                & CLIP-Guided         & 28.4 & 5.5 & 2.96 & 0.49 & 66.1\\
                                \rowcolor{gray!30}
\cellcolor{white}               & \ours-Guided        & 26.2 & 4.6 & 3.30 & 0.39 & 66.8\\
\midrule
\multirow{3}{*}{LLaVA-OV-7B}   & Greedy (Default)    & 35.0 & 5.5 & 3.12 & 0.36 & 74.5 \\
                                & CLIP-Guided        & 29.4 & 4.8 & 3.27 & 0.34 & 75.5 \\
                                \rowcolor{gray!30}
\cellcolor{white}               & \textbf{\ours-Guided}        & 27.0 & 3.7 & 3.34 & 0.31 & 76.2 \\
\midrule
\multirow{3}{*}{Qwen2-VL-7B}    & Greedy (Default)    & 30.8 & 5.2 & 3.27 & 0.37 & 69.4 \\
                                & CLIP-Guided         & 27.3 & 3.9 & 3.33 & 0.32 & 70.2 \\
                                \rowcolor{gray!30}
\cellcolor{white}               & \textbf{\ours-Guided}        & 24.5 & 3.3 & 3.39 & 0.29 & 73.5 \\
\bottomrule
\end{tabular}
}
\vspace{-10pt}
\end{table} 

\subsubsection*{\circled{2} \ours-Guided Search Reduces Visual Hallucination}
\label{sec: 4.1.2}

To benchmark the benefits of \ours in improving visual comprehension, we evaluate the degree of visual hallucination present in the generated responses. Following the setting in previous works~\cite{zhou2024aligning, zhou2024calibrated}, we randomly sample 500 images from the COCO Val2014 dataset and use prompts from the LLaVA-150k detailed description dataset.
The widely used CHAIR~\cite{rohrbach2018object} metric is used for hallucination evaluation
and we also use MMHal~\cite{sun2023aligninglargemultimodalmodels} as another benchmark for hallucination evaluation.
Besides, we adopt the coverage metric from AMBER~\cite{wang2024amberllmfreemultidimensionalbenchmark} to evaluate the object coverage of generated captions, thus preventing artificially low hallucination scores caused by overly short captions 

The experiment results based on LLaVA-Next-7B in Table~\ref{tab: infer_eval} show that \ours-guided search significantly outperforms greedy decoding, BoN, and CLIP-guided search, reducing CHAIRs from $32.4$ to $26.2$,  CHAIRi from $5.9$ to $4.6$, MMHal rate from $0.52$ to $0.39$, and improving MMHal from $2.94$ to $3.30$. 
Meanwhile, object coverage improves from 63.9 to 66.8, indicating that the reduced hallucination brought by \ours is not through generating short captions.
Compared to MCTS, \ours achieves comparable or superior performance while requiring significantly lower computation cost, highlighting its efficiency and effectiveness.

The reduction in hallucination within the image descriptions generated via \ours-guided search aligns with our training objective for \ours. Specifically, using the CLIP score as a reward, \ours is trained through TD learning to select responses at each step that minimize future hallucinations, thereby enhancing the overall response quality.

To validate the robustness of \ours, we retrain the corresponding \ours based on LLaVA-OV-7B and Qwen2-VL-7B-Instruct, following the procedure in Section~\ref{sec 3.2}. 
Table~\ref{tab: infer_eval}'s results indicate that \ours can effectively mitigate hallucinations even when applied to stronger VLMs.

\subsubsection*{\circled{3} Benefits from Further Scaling Up Inference Compute}

We next investigate the impact of scaling up the inference-time compute on the VLM response quality at each step, by changing the search step sizes.  
To support a larger maximum step size, we only keep $T=0.5$ as the temperature configuration when experimenting with different step sizes.
We use CHAIRs as the evaluation metric, with the same evaluation data and prompts as in Table~\ref{tab: infer_eval}. 
We report the CHAIRs scores for image descriptions obtained using \ours-guided search and CLIP-PRM-guided search at step sizes of 2, 4, 8, and 16. 
The experimental results are depicted in Figure~\ref{fig:scaling_curve}.

We observe that the performance of both \ours-guided search and CLIP-PRM-guided search improves progressively as the search step size increases, indicating that scaling inference-time computation can enhance the performance of VLMs. 
Notably, as the step size grows, the performance improvement of \ours-guided search accelerates at a faster rate, resulting in a widening performance gap between the two methods. Additionally, \ours proves to be nearly twice as computationally efficient as CLIP-PRM for reaching comparable performance: at a step size of 8, \ours achieves results comparable to those of CLIP-PRM at a step size of 16. These findings further validate the effectiveness and efficiency of \ours as a superior inference-time search signal for VLMs.

\subsubsection*{\circled{4} Stronger PRM can Further Enhance \ours}

In the previous and next sections, motivated by self-improvement, we consistently select the visual encoder corresponding to the base VLMs as PRM for VisVM training. 
In this subsection, we conduct an ablation study to demonstrate the generality of the VisVM training pipeline. 
Specifically, we utilize a more powerful model, SigLIP, as the PRM to train VisVM, while maintaining LLaVA-Next-7B as the base model. 
The remaining training procedures are identical to those used when CLIP served as the PRM. We evaluate the performance of VisVM trained with different PRMs using the CHAIR, MMHal, and AMBER Cov metrics; the results are presented in Table~\ref{tab: prm_ablation}. 
\begin{table}[t]
\centering
\caption{Ablation study of different PRMs for \ours training. We observe that stronger PRM lead to better \ours performance.}
\label{tab: prm_ablation}
\tablestyle{8pt}{1.1} 
\resizebox{\linewidth}{!}{%
\begin{tabular}{l|cccccc}
\toprule 
Searching Method & CHAIRs $\downarrow$ & CHAIRi $\downarrow$ & MMHal $\uparrow$ & MMHal rate $\downarrow$ & AMBER Cov $\uparrow$\\
\midrule
Greedy (Default)             & 32.4 & 5.9 & 2.94 & 0.52 & 63.9\\
CLIP-\ours-Guided            & 26.2 & 4.6 & 3.30 & 0.39 & 66.8 \\
SigLIP-\ours-Guided          & 25.6 & 4.4 & 3.31 & 0.36 & 67.5 \\
\bottomrule
\end{tabular}
}
\vspace{-10pt}
\end{table} 
Notably, using SigLIP as the PRM results in significantly reduced hallucinations in captions generated through VisVM guided search, with clear improvements observed particularly in CHAIR and AMBER Cov scores. 
This finding indicates that leveraging a stronger PRM further enhances VisVM capabilities, underscoring the generalizability and strong potential of the VisVM training framework.

\definecolor{front-color}{HTML}{F5FFFA}
\begin{table*}
\centering
\caption{Performance after fine-tuning LLAVA-Next-Mistral-7B and Qwen2-VL-7B-Instruct with image descriptions obtained using different search methods. The model with \ours search as data source achieves the best performance across all benchmarks, with an average improvement of 10.8\% and 7.3\% compared with the base model, respectively. We calculate the final performance improvement using 100-CHAIRs, 10-CHAIRi, and 1-MMHal rate respectively.}
\label{tab: self-training}
\tablestyle{9pt}{1.1} 
\resizebox{\linewidth}{!}{%

\begin{tabular}{l|l|ccccccc|cccc|c}
\toprule 
 & & \multicolumn{7}{c|}{Visual Comprehension Benchmark} & \multicolumn{4}{c|}{Hallucination Benchmark}  \\
\cmidrule(lr){3-9}\cmidrule(lr){10-13}
Base & SFT Data Source & \rotatebox{90}{MM-Vet} $\uparrow$ & \rotatebox{90}{MMBench} $\uparrow$ & \rotatebox{90}{MMMU}  $\uparrow$ & \rotatebox{90}{MathVista}  $\uparrow$ & \rotatebox{90}{CVBench} $\uparrow$ & \rotatebox{90}{LLAVA$^{\mathrm{W}}$} $\uparrow$ & \rotatebox{90}{MMStar} $\uparrow$ & \rotatebox{90}{CHAIRs} $\downarrow$ & \rotatebox{90}{CHAIRi} $\downarrow$ & \rotatebox{90}{MMHal} $\uparrow$ & \rotatebox{90}{MMHal} \rotatebox{90}{rate} $\downarrow$ & Avg.\\
\midrule
\multirow{5}{*}{LLaVA-Next-7B}  & --                 & 45.2 & 74.9 & 34.2 & 38.5 & 65.8 & 76.9 & 36.0 & 32.4 & 5.9 & 2.94 & 0.52 & --\\
                                & Greedy decoding    & 43.5 & 74.6 & 34.9 & 37.8 & 66.2 & 75.1 & 36.7 & 33.2 & 6.3 & 2.97 & 0.54 & -1.6\%\\
                                & GPT4o-BoN (30)     & 47.1 & 76.1 & 35.4 & 40.9 & 67.9 & 77.3 & 36.9 & 30.0 & 5.4 & 3.11 & 0.47 & +4.9\%\\
                                & CLIP-PRM search    & 46.1 & 75.8 & 35.8 & 39.6 & 68.5 & 78.1 & 36.6 & 26.0 & 5.2 & 3.01 & 0.50 & +4.6\%\\
                                \rowcolor{gray!30}
\cellcolor{white}               & \ours search    & \textbf{48.3} & \textbf{76.7} & \textbf{36.1} & \textbf{42.3}  & \textbf{69.8} & \textbf{78.4} & \textbf{38.0} & \textbf{22.6} & \textbf{4.3} & \textbf{3.26} & \textbf{0.44} & \textbf{\textcolor[HTML]{548235}{+10.8\%}}\\
\midrule
\multirow{5}{*}{Qwen2-VL-7B}  & --                   & 58.4 & 83.0 & 49.3 & 58.2 & 74.5 & 87.1 & 56.3 & 30.8 & 5.2 & 3.27 & 0.37 & --\\
                                & Greedy decoding    & 58.3 & 83.1 & 49.4 & 58.7 & 74.1 & 86.3 & 56.5 & 29.7 & 5.1 & 3.13 & 0.42 & -0.8\%\\
                                & GPT4o-BoN (30)     & 58.8 & 83.7 & 49.3 & 60.2 & 74.6 & 87.2 & 56.7 & 25.4 & 4.0 & 3.31 & 0.35 & +3.9\%\\
                                & CLIP-PRM search    & 58.5 & 83.5 & 49.5 & 59.2 & 74.9 & 87.9 & 56.5 & 23.6 & 3.7 & 3.31 & 0.32 & +5.0\%\\
                                \rowcolor{gray!30}
\cellcolor{white}             & \ours search         & \textbf{58.9} & \textbf{84.1} & \textbf{49.7} & \textbf{61.1}  & \textbf{76.2} & \textbf{88.2} & \textbf{57.0} & \textbf{21.4} & \textbf{3.4} & \textbf{3.34} & \textbf{0.28} & \textbf{\textcolor[HTML]{548235}{+7.3\%}}\\
\bottomrule
\end{tabular}
}
\vspace{-10pt}
\end{table*}

\subsection{Self-Training Vision-Language Model}
\label{sec_42}

Inference-time search with \ours proves to be an effective approach in boosting VLMs' visual comprehension capability. This naturally motivates the question:  Can we use the higher-quality descriptive captions generated by \ours-guided search to further improve the original VLM, thereby enabling a form of self-training pipeline?

\paragraph{Training details.}
We start with the 9,215 $<$image, prompt$>$ pairs from Section~\ref{sec 3.2}, which are used to generate VisVM training data. 
To demonstrate the robustness of our method, we conduct experiments using two different VLMs, \textbf{LLaVA-Next-Mistral-7B} and \textbf{Qwen2-VL-7B-Instruct}, as the base models.
We first generate corresponding image descriptions for all 9,215 $<$image, prompt$>$ pairs using \ours-guided search, resulting in 9,215 $<$image, prompt, description$>$ tuples as the SFT dataset. 
Subsequently, we conduct a full parameter fine-tuning on base VLMs using this SFT dataset for three epochs with a learning rate of 1e-6. 
As a comparison, we also generate corresponding descriptions on this prompt dataset using greedy decoding, BoN, and CLIP-PRM-guided search, and perform full parameter SFT on base models with the same learning rate and number of epochs.
All experiments are conducted on 8$\times$80GB A100 GPUs.

\paragraph{Evaluation benchmarks.}
We conduct evaluations on two types of benchmarks: visual comprehension benchmarks and hallucination benchmarks.
For the visual comprehension evaluation, we select seven standard benchmarks:
MM-Vet~\cite{yu2023mmvetevaluatinglargemultimodal}, MMBench~\cite{liu2024mmbenchmultimodalmodelallaround}, MMMU~\cite{yue2024mmmumassivemultidisciplinemultimodal}, MathVista~\cite{lu2024mathvista}, CVBench~\cite{tong2024cambrian1fullyopenvisioncentric}, LLAVA-Wild~\cite{liu2023visualinstructiontuning}, and MMStar~\cite{chen2024rightwayevaluatinglarge}.
For hallucination evaluation, we benchmark on CHAIR~\cite{rohrbach2018object} and MMHal~\cite{sun2023aligninglargemultimodalmodels}.

\paragraph{Evaluation results on visual comprehension.} 
Table~\ref{tab: self-training} presents the fine-tuning results of LLaVA-Next and Qwen2-VL on visual comprehension benchmarks. 
Performance improved across nearly all benchmarks after self-training, with one exception of the greedy decoding self-training, which leads to a decline in most cases. Among the methods evaluated, the \ours search self-training approach demonstrates the most significant improvement, boosting LLaVA-Next and Qwen2-VL average performance by \textbf{5.5\%} and \textbf{1.8\%}, respectively. This gain far exceeds the improvements achieved by the BoN and CLIP-PRM search methods. These findings highlight the superior quality of descriptive captions obtained through \ours search, which significantly enhances VLM’s visual comprehension capabilities during self-training.

\paragraph{Evaluation results on visual hallucinations.} 
As shown in Table~\ref{tab: self-training}, the \ours search self-training can also significantly reduce hallucination in VLM. When evaluated across four metrics on two benchmarks, \ours search self-training decreases the hallucination rates of LLaVA-Next and Qwen2-VL by 20.3\% and 16.9\%, substantially outperforming the reductions achieved by BoN and CLIP-PRM search. These results further validate the effectiveness of the \ours search self-training approach.

\paragraph{The promise of a VLM self-training pipeline.} 
The experiment results in this section demonstrate that the \ours search significantly enhances the visual comprehension capabilities of LLaVA-Next and Qwen2-VL by generating high-quality descriptive captions as the SFT data. Throughout this process, no external models or human annotations are utilized beyond the raw COCO images. The reward model for training \ours is derived from the visual encoder embedded within LLaVA-Next and Qwen2-VL, and \ours itself is initialized from the parameters of LLaVA-Next and Qwen2-VL. The SFT data is produced by \ours-guided search using base VLMs, ensuring that all training signals originated solely from the same VLM. 
As future directions, we see great promise in applying this method to other VLMs, leading to a genuine self-training pipeline that could continuously self-improve VLMs' visual comprehension capability, without reliance on any external models or human annotations.

\subsection{\ours Analysis}
\begin{table}[t]
    \centering
    \caption{Hallucination comparison of VisVM and CLIP selection starting from same sentence candidates.}
    \label{tab: visvm_analyse}
    \resizebox{0.6\linewidth}{!}{%
    \begin{tabular}{l|cc}
        \toprule
        Selection Model & CHAIRs $\downarrow$ & CHAIRi $\downarrow$\\
        \midrule
        CLIP              & 31.6 & 5.7\\
        VisVM             & 30.9 & 5.3\\
        \bottomrule
    \end{tabular}}
    \vspace{-6pt}
\end{table}
To further understand how \ours enhances response quality by predicting future values, we design a quantitative experiment in this section to compare the effects on image captioning when selecting step candidates using \ours versus CLIP.
We follow the experimental settings described in Section~\ref{sec: 4.1.2}, randomly sampling 500 images from the COCO Val2014 dataset and using prompts from the LLaVA-150k detailed description dataset. For each $<$image, prompt$>$ pair, we employ the LLaVA-Next-7B model to generate six candidate sentences, including greedy decoding and five different temperature settings. 
Subsequently, we select one candidate sentence from these six candidates using the \ours and CLIP models independently. 
We then utilize the LLaVA-Next-7B model to continue generating a complete image description via greedy decoding based on the selected candidate sentence. 
Finally, we evaluate hallucinations within generated descriptions using the CHAIR metric, with results shown in Table~\ref{tab: visvm_analyse}.

Despite the selection being made from the same set of sentence candidates, differences arise in the selected candidates due to \ours's ability to predict long-term value, resulting in fewer hallucinations in captions generated by greedy decoding. In \ours-guided search, this predictive selection by \ours is applied at each step, significantly minimizing the occurrence of hallucinations in the final response. We provide a more detailed case study in Appendix~\ref{app: Analysis} to further illustrate this.

\section{Conclusion}
\label{sec:Conclusion}
We have presented \ours, a vision value model that effectively guides VLM for inference-time search to improve visual comprehension. Our results demonstrate that scaling inference-time computations can produce VLM responses that include richer visual details and reduce hallucinations. Among various reward signals, \ours has a better scaling behavior due to its consideration of potential future generations. Moreover, we highlight the promise of using \ours-guided search to establish a self-training pipeline, enabling the enhancement of VLMs without external annotations.

\section*{Acknowledgment}
Wang, Xu, and Huang are supported by DARPA Transfer from Imprecise and Abstract Models to Autonomous Technologies (TIAMAT) 80321, DARPA HR001124S0029-AIQ-FP-019, DOD-AFOSR-Air Force Office of Scientific Research under award number FA9550-23-1-0048, National Science Foundation NSF-IIS-2147276 FAI, National Science Foundation NAIRR240045, National Science Foundation TRAILS Institute (2229885). Private support was provided by Peraton.

{
    \small
    \bibliographystyle{ieeenat_fullname}
    \bibliography{main}

\begin{thebibliography}{75}
\providecommand{\natexlab}[1]{#1}
\providecommand{\url}[1]{\texttt{#1}}
\expandafter\ifx\csname urlstyle\endcsname\relax
  \providecommand{\doi}[1]{doi: #1}\else
  \providecommand{\doi}{doi: \begingroup \urlstyle{rm}\Url}\fi

\bibitem[o1b(2024)]{o1blog}
Learning to reason with llms.
\newblock \url{https://openai.com/index/learning-to-reason-with-llms/}, 2024.

\bibitem[Alayrac et~al.(2022)Alayrac, Donahue, Luc, Miech, Barr, Hasson, Lenc, Mensch, Millican, Reynolds, et~al.]{alayrac2022flamingo}
Jean-Baptiste Alayrac, Jeff Donahue, Pauline Luc, Antoine Miech, Iain Barr, Yana Hasson, Karel Lenc, Arthur Mensch, Katherine Millican, Malcolm Reynolds, et~al.
\newblock Flamingo: a visual language model for few-shot learning.
\newblock \emph{Advances in Neural Information Processing Systems}, 35:\penalty0 23716--23736, 2022.

\bibitem[Bai et~al.(2023)Bai, Bai, Yang, Wang, Tan, Wang, Lin, Zhou, and Zhou]{bai2023qwen}
Jinze Bai, Shuai Bai, Shusheng Yang, Shijie Wang, Sinan Tan, Peng Wang, Junyang Lin, Chang Zhou, and Jingren Zhou.
\newblock Qwen-vl: A frontier large vision-language model with versatile abilities.
\newblock \emph{arXiv preprint arXiv:2308.12966}, 2023.

\bibitem[Betker et~al.(2023)Betker, Goh, Jing, Brooks, Wang, Li, Ouyang, Zhuang, Lee, Guo, et~al.]{betker2023improving}
James Betker, Gabriel Goh, Li Jing, Tim Brooks, Jianfeng Wang, Linjie Li, Long Ouyang, Juntang Zhuang, Joyce Lee, Yufei Guo, et~al.
\newblock Improving image generation with better captions.
\newblock \emph{Computer Science. https://cdn. openai. com/papers/dall-e-3. pdf}, 2\penalty0 (3):\penalty0 8, 2023.

\bibitem[Brown et~al.(2024)Brown, Juravsky, Ehrlich, Clark, Le, R{\'e}, and Mirhoseini]{brown2024large}
Bradley Brown, Jordan Juravsky, Ryan Ehrlich, Ronald Clark, Quoc~V Le, Christopher R{\'e}, and Azalia Mirhoseini.
\newblock Large language monkeys: Scaling inference compute with repeated sampling.
\newblock \emph{arXiv preprint arXiv:2407.21787}, 2024.

\bibitem[Chakraborty et~al.(2024)Chakraborty, Ghosal, Yin, Manocha, Wang, Bedi, and Huang]{chakraborty2024transfer}
Souradip Chakraborty, Soumya~Suvra Ghosal, Ming Yin, Dinesh Manocha, Mengdi Wang, Amrit~Singh Bedi, and Furong Huang.
\newblock Transfer q star: Principled decoding for llm alignment.
\newblock \emph{arXiv preprint arXiv:2405.20495}, 2024.

\bibitem[Changpinyo et~al.(2021)Changpinyo, Sharma, Ding, and Soricut]{changpinyo2021conceptual}
Soravit Changpinyo, Piyush Sharma, Nan Ding, and Radu Soricut.
\newblock Conceptual 12m: Pushing web-scale image-text pre-training to recognize long-tail visual concepts.
\newblock In \emph{Proceedings of the IEEE/CVF Conference on Computer Vision and Pattern Recognition}, pages 3558--3568, 2021.

\bibitem[Chen et~al.(2023{\natexlab{a}})Chen, Li, Dong, Zhang, He, Wang, Zhao, and Lin]{chen2023sharegpt4v}
Lin Chen, Jinsong Li, Xiaoyi Dong, Pan Zhang, Conghui He, Jiaqi Wang, Feng Zhao, and Dahua Lin.
\newblock Sharegpt4v: Improving large multi-modal models with better captions.
\newblock \emph{arXiv preprint arXiv:2311.12793}, 2023{\natexlab{a}}.

\bibitem[Chen et~al.(2024{\natexlab{a}})Chen, Li, Dong, Zhang, Zang, Chen, Duan, Wang, Qiao, Lin, and Zhao]{chen2024rightwayevaluatinglarge}
Lin Chen, Jinsong Li, Xiaoyi Dong, Pan Zhang, Yuhang Zang, Zehui Chen, Haodong Duan, Jiaqi Wang, Yu Qiao, Dahua Lin, and Feng Zhao.
\newblock Are we on the right way for evaluating large vision-language models?, 2024{\natexlab{a}}.

\bibitem[Chen et~al.(2015)Chen, Fang, Lin, Vedantam, Gupta, Doll{\'a}r, and Zitnick]{chen2015microsoft}
Xinlei Chen, Hao Fang, Tsung-Yi Lin, Ramakrishna Vedantam, Saurabh Gupta, Piotr Doll{\'a}r, and C~Lawrence Zitnick.
\newblock Microsoft coco captions: Data collection and evaluation server.
\newblock \emph{arXiv preprint arXiv:1504.00325}, 2015.

\bibitem[Chen et~al.(2023{\natexlab{b}})Chen, Zhu, Zhan, Li, Zhao, Wang, and Tang]{chen2023mitigatinghallucinationvisuallanguage}
Zhiyang Chen, Yousong Zhu, Yufei Zhan, Zhaowen Li, Chaoyang Zhao, Jinqiao Wang, and Ming Tang.
\newblock Mitigating hallucination in visual language models with visual supervision, 2023{\natexlab{b}}.

\bibitem[Chen et~al.(2024{\natexlab{b}})Chen, Wu, Wang, Su, Chen, Xing, Zhong, Zhang, Zhu, Lu, et~al.]{chen2024internvl}
Zhe Chen, Jiannan Wu, Wenhai Wang, Weijie Su, Guo Chen, Sen Xing, Muyan Zhong, Qinglong Zhang, Xizhou Zhu, Lewei Lu, et~al.
\newblock Internvl: Scaling up vision foundation models and aligning for generic visual-linguistic tasks.
\newblock In \emph{Proceedings of the IEEE/CVF Conference on Computer Vision and Pattern Recognition}, pages 24185--24198, 2024{\natexlab{b}}.

\bibitem[Cobbe et~al.(2021)Cobbe, Kosaraju, Bavarian, Chen, Jun, Kaiser, Plappert, Tworek, Hilton, Nakano, et~al.]{cobbe2021training}
Karl Cobbe, Vineet Kosaraju, Mohammad Bavarian, Mark Chen, Heewoo Jun, Lukasz Kaiser, Matthias Plappert, Jerry Tworek, Jacob Hilton, Reiichiro Nakano, et~al.
\newblock Training verifiers to solve math word problems.
\newblock \emph{arXiv preprint arXiv:2110.14168}, 2021.

\bibitem[Esser et~al.(2024)Esser, Kulal, Blattmann, Entezari, M{\"u}ller, Saini, Levi, Lorenz, Sauer, Boesel, et~al.]{esser2024scaling}
Patrick Esser, Sumith Kulal, Andreas Blattmann, Rahim Entezari, Jonas M{\"u}ller, Harry Saini, Yam Levi, Dominik Lorenz, Axel Sauer, Frederic Boesel, et~al.
\newblock Scaling rectified flow transformers for high-resolution image synthesis.
\newblock In \emph{Forty-first International Conference on Machine Learning}, 2024.

\bibitem[Goyal et~al.(2017)Goyal, Khot, Summers-Stay, Batra, and Parikh]{vqav2}
Yash Goyal, Tejas Khot, Douglas Summers-Stay, Dhruv Batra, and Devi Parikh.
\newblock Making the v in vqa matter: Elevating the role of image understanding in visual question answering.
\newblock In \emph{Proceedings of the IEEE conference on computer vision and pattern recognition}, pages 6904--6913, 2017.

\bibitem[Guan et~al.(2024)Guan, Liu, Wu, Xian, Li, Liu, Wang, Chen, Huang, Yacoob, Manocha, and Zhou]{guan2024hallusionbenchadvanceddiagnosticsuite}
Tianrui Guan, Fuxiao Liu, Xiyang Wu, Ruiqi Xian, Zongxia Li, Xiaoyu Liu, Xijun Wang, Lichang Chen, Furong Huang, Yaser Yacoob, Dinesh Manocha, and Tianyi Zhou.
\newblock Hallusionbench: An advanced diagnostic suite for entangled language hallucination and visual illusion in large vision-language models, 2024.

\bibitem[Hansen et~al.(2022)Hansen, Wang, and Su]{hansen2022temporaldifferencelearningmodel}
Nicklas Hansen, Xiaolong Wang, and Hao Su.
\newblock Temporal difference learning for model predictive control, 2022.

\bibitem[Hosseini et~al.(2024)Hosseini, Yuan, Malkin, Courville, Sordoni, and Agarwal]{hosseini2024vstartrainingverifiersselftaught}
Arian Hosseini, Xingdi Yuan, Nikolay Malkin, Aaron Courville, Alessandro Sordoni, and Rishabh Agarwal.
\newblock V-star: Training verifiers for self-taught reasoners, 2024.

\bibitem[Hu et~al.(2022)Hu, Gan, Wang, Yang, Liu, Lu, and Wang]{hu2021scaling}
Xiaowei Hu, Zhe Gan, Jianfeng Wang, Zhengyuan Yang, Zicheng Liu, Yumao Lu, and Lijuan Wang.
\newblock Scaling up vision-language pre-training for image captioning.
\newblock In \emph{CVPR}, 2022.

\bibitem[Jia et~al.(2021)Jia, Yang, Xia, Chen, Parekh, Pham, Le, Sung, Li, and Duerig]{jia2021scaling}
Chao Jia, Yinfei Yang, Ye Xia, Yi-Ting Chen, Zarana Parekh, Hieu Pham, Quoc~V Le, Yunhsuan Sung, Zhen Li, and Tom Duerig.
\newblock Scaling up visual and vision-language representation learning with noisy text supervision.
\newblock \emph{arXiv preprint arXiv:2102.05918}, 2021.

\bibitem[Jiang et~al.(2024)Jiang, He, Zeng, Wei, Ku, Liu, and Chen]{jiang2024mantisinterleavedmultiimageinstruction}
Dongfu Jiang, Xuan He, Huaye Zeng, Cong Wei, Max Ku, Qian Liu, and Wenhu Chen.
\newblock Mantis: Interleaved multi-image instruction tuning, 2024.

\bibitem[Lai et~al.(2024)Lai, Saveris, Chen, Chen, Zhang, Zhang, Tebar, Hu, Gan, Grasch, et~al.]{lai2024revisit}
Zhengfeng Lai, Vasileios Saveris, Chen Chen, Hong-You Chen, Haotian Zhang, Bowen Zhang, Juan~Lao Tebar, Wenze Hu, Zhe Gan, Peter Grasch, et~al.
\newblock Revisit large-scale image-caption data in pre-training multimodal foundation models.
\newblock \emph{arXiv preprint arXiv:2410.02740}, 2024.

\bibitem[Lai et~al.(2025)Lai, Zhang, Zhang, Wu, Bai, Timofeev, Du, Gan, Shan, Chuah, et~al.]{lai2025veclip}
Zhengfeng Lai, Haotian Zhang, Bowen Zhang, Wentao Wu, Haoping Bai, Aleksei Timofeev, Xianzhi Du, Zhe Gan, Jiulong Shan, Chen-Nee Chuah, et~al.
\newblock Veclip: Improving clip training via visual-enriched captions.
\newblock In \emph{European Conference on Computer Vision}, pages 111--127. Springer, 2025.

\bibitem[Leng et~al.(2023)Leng, Zhang, Chen, Li, Lu, Miao, and Bing]{leng2023mitigatingobjecthallucinationslarge}
Sicong Leng, Hang Zhang, Guanzheng Chen, Xin Li, Shijian Lu, Chunyan Miao, and Lidong Bing.
\newblock Mitigating object hallucinations in large vision-language models through visual contrastive decoding, 2023.

\bibitem[Li et~al.(2024{\natexlab{a}})Li, Zhang, Guo, Zhang, Li, Zhang, Zhang, Zhang, Li, Liu, and Li]{li2024llavaonevisioneasyvisualtask}
Bo Li, Yuanhan Zhang, Dong Guo, Renrui Zhang, Feng Li, Hao Zhang, Kaichen Zhang, Peiyuan Zhang, Yanwei Li, Ziwei Liu, and Chunyuan Li.
\newblock Llava-onevision: Easy visual task transfer, 2024{\natexlab{a}}.

\bibitem[Li et~al.(2024{\natexlab{b}})Li, Gan, Yang, Yang, Li, Wang, Gao, et~al.]{li2024multimodal}
Chunyuan Li, Zhe Gan, Zhengyuan Yang, Jianwei Yang, Linjie Li, Lijuan Wang, Jianfeng Gao, et~al.
\newblock Multimodal foundation models: From specialists to general-purpose assistants.
\newblock \emph{Foundations and Trends{\textregistered} in Computer Graphics and Vision}, 16\penalty0 (1-2):\penalty0 1--214, 2024{\natexlab{b}}.

\bibitem[Li et~al.(2024{\natexlab{c}})Li, Wang, Hu, Wei, Zheng, Hu, Zhang, and Peng]{li2024common7blanguagemodels}
Chen Li, Weiqi Wang, Jingcheng Hu, Yixuan Wei, Nanning Zheng, Han Hu, Zheng Zhang, and Houwen Peng.
\newblock Common 7b language models already possess strong math capabilities, 2024{\natexlab{c}}.

\bibitem[Li et~al.(2024{\natexlab{d}})Li, Zhang, Zhang, Zhang, Li, Li, Ma, and Li]{li2024llavanextinterleavetacklingmultiimagevideo}
Feng Li, Renrui Zhang, Hao Zhang, Yuanhan Zhang, Bo Li, Wei Li, Zejun Ma, and Chunyuan Li.
\newblock Llava-next-interleave: Tackling multi-image, video, and 3d in large multimodal models, 2024{\natexlab{d}}.

\bibitem[Lightman et~al.(2023{\natexlab{a}})Lightman, Kosaraju, Burda, Edwards, Baker, Lee, Leike, Schulman, Sutskever, and Cobbe]{lightman2023let}
Hunter Lightman, Vineet Kosaraju, Yura Burda, Harri Edwards, Bowen Baker, Teddy Lee, Jan Leike, John Schulman, Ilya Sutskever, and Karl Cobbe.
\newblock Let's verify step by step.
\newblock \emph{arXiv preprint arXiv:2305.20050}, 2023{\natexlab{a}}.

\bibitem[Lightman et~al.(2023{\natexlab{b}})Lightman, Kosaraju, Burda, Edwards, Baker, Lee, Leike, Schulman, Sutskever, and Cobbe]{lightman2023letsverifystepstep}
Hunter Lightman, Vineet Kosaraju, Yura Burda, Harri Edwards, Bowen Baker, Teddy Lee, Jan Leike, John Schulman, Ilya Sutskever, and Karl Cobbe.
\newblock Let's verify step by step, 2023{\natexlab{b}}.

\bibitem[Liu et~al.(2023{\natexlab{a}})Liu, Lin, Li, Wang, Yacoob, and Wang]{liu2023aligning}
Fuxiao Liu, Kevin Lin, Linjie Li, Jianfeng Wang, Yaser Yacoob, and Lijuan Wang.
\newblock Aligning large multi-modal model with robust instruction tuning.
\newblock \emph{arXiv preprint arXiv:2306.14565}, 2023{\natexlab{a}}.

\bibitem[Liu et~al.(2023{\natexlab{b}})Liu, Li, Wu, and Lee]{liu2023visualinstructiontuning}
Haotian Liu, Chunyuan Li, Qingyang Wu, and Yong~Jae Lee.
\newblock Visual instruction tuning, 2023{\natexlab{b}}.

\bibitem[Liu et~al.(2023{\natexlab{c}})Liu, Li, Wu, and Lee]{llava}
Haotian Liu, Chunyuan Li, Qingyang Wu, and Yong~Jae Lee.
\newblock Visual instruction tuning.
\newblock \emph{arXiv preprint arXiv:2304.08485}, 2023{\natexlab{c}}.

\bibitem[Liu et~al.(2024)Liu, Duan, Zhang, Li, Zhang, Zhao, Yuan, Wang, He, Liu, Chen, and Lin]{liu2024mmbenchmultimodalmodelallaround}
Yuan Liu, Haodong Duan, Yuanhan Zhang, Bo Li, Songyang Zhang, Wangbo Zhao, Yike Yuan, Jiaqi Wang, Conghui He, Ziwei Liu, Kai Chen, and Dahua Lin.
\newblock Mmbench: Is your multi-modal model an all-around player?, 2024.

\bibitem[Lu et~al.(2024)Lu, Bansal, Xia, Liu, Li, Hajishirzi, Cheng, Chang, Galley, and Gao]{lu2024mathvista}
Pan Lu, Hritik Bansal, Tony Xia, Jiacheng Liu, Chunyuan Li, Hannaneh Hajishirzi, Hao Cheng, Kai-Wei Chang, Michel Galley, and Jianfeng Gao.
\newblock Mathvista: Evaluating mathematical reasoning of foundation models in visual contexts.
\newblock In \emph{International Conference on Learning Representations (ICLR)}, 2024.

\bibitem[OpenAI(2023)]{gpt4v}
OpenAI.
\newblock Gpt-4v(ision) system card.
\newblock 2023.

\bibitem[Radford et~al.(2021)Radford, Kim, Hallacy, Ramesh, Goh, Agarwal, Sastry, Askell, Mishkin, Clark, et~al.]{radford2021learning}
Alec Radford, Jong~Wook Kim, Chris Hallacy, Aditya Ramesh, Gabriel Goh, Sandhini Agarwal, Girish Sastry, Amanda Askell, Pamela Mishkin, Jack Clark, et~al.
\newblock Learning transferable visual models from natural language supervision.
\newblock \emph{arXiv preprint arXiv:2103.00020}, 2021.

\bibitem[Rohrbach et~al.(2018)Rohrbach, Hendricks, Burns, Darrell, and Saenko]{rohrbach2018object}
Anna Rohrbach, Lisa~Anne Hendricks, Kaylee Burns, Trevor Darrell, and Kate Saenko.
\newblock Object hallucination in image captioning.
\newblock \emph{arXiv preprint arXiv:1809.02156}, 2018.

\bibitem[Shi et~al.(2024)Shi, Liu, Wang, Liao, Radhakrishnan, Huang, Yin, Sapra, Yacoob, Shi, et~al.]{shi2024eagle}
Min Shi, Fuxiao Liu, Shihao Wang, Shijia Liao, Subhashree Radhakrishnan, De-An Huang, Hongxu Yin, Karan Sapra, Yaser Yacoob, Humphrey Shi, et~al.
\newblock Eagle: Exploring the design space for multimodal llms with mixture of encoders.
\newblock \emph{arXiv preprint arXiv:2408.15998}, 2024.

\bibitem[Silver et~al.(2016)Silver, Huang, Maddison, Guez, Sifre, Van Den~Driessche, Schrittwieser, Antonoglou, Panneershelvam, Lanctot, et~al.]{silver2016mastering}
David Silver, Aja Huang, Chris~J Maddison, Arthur Guez, Laurent Sifre, George Van Den~Driessche, Julian Schrittwieser, Ioannis Antonoglou, Veda Panneershelvam, Marc Lanctot, et~al.
\newblock Mastering the game of go with deep neural networks and tree search.
\newblock \emph{nature}, 529\penalty0 (7587):\penalty0 484--489, 2016.

\bibitem[Snell et~al.(2024{\natexlab{a}})Snell, Lee, Xu, and Kumar]{snell2024scaling}
Charlie Snell, Jaehoon Lee, Kelvin Xu, and Aviral Kumar.
\newblock Scaling llm test-time compute optimally can be more effective than scaling model parameters.
\newblock \emph{arXiv preprint arXiv:2408.03314}, 2024{\natexlab{a}}.

\bibitem[Snell et~al.(2024{\natexlab{b}})Snell, Lee, Xu, and Kumar]{snell2024scalingllmtesttimecompute}
Charlie Snell, Jaehoon Lee, Kelvin Xu, and Aviral Kumar.
\newblock Scaling llm test-time compute optimally can be more effective than scaling model parameters, 2024{\natexlab{b}}.

\bibitem[Sun et~al.(2023)Sun, Shen, Cao, Liu, Li, Shen, Gan, Gui, Wang, Yang, Keutzer, and Darrell]{sun2023aligninglargemultimodalmodels}
Zhiqing Sun, Sheng Shen, Shengcao Cao, Haotian Liu, Chunyuan Li, Yikang Shen, Chuang Gan, Liang-Yan Gui, Yu-Xiong Wang, Yiming Yang, Kurt Keutzer, and Trevor Darrell.
\newblock Aligning large multimodal models with factually augmented rlhf, 2023.

\bibitem[Sutton(1988)]{sutton1988learning}
Richard~S Sutton.
\newblock Learning to predict by the methods of temporal differences.
\newblock \emph{Machine learning}, 3:\penalty0 9--44, 1988.

\bibitem[Team(2023)]{geminiteam2023gemini}
Google~Gemini Team.
\newblock Gemini: A family of highly capable multimodal models, 2023.

\bibitem[Teng et~al.(2023)Teng, Hu, Deng, Li, Li, Ai, Yang, Li, Xuanyuan, Zhu, and Chen]{Teng_2023}
Siyu Teng, Xuemin Hu, Peng Deng, Bai Li, Yuchen Li, Yunfeng Ai, Dongsheng Yang, Lingxi Li, Zhe Xuanyuan, Fenghua Zhu, and Long Chen.
\newblock Motion planning for autonomous driving: The state of the art and future perspectives.
\newblock \emph{IEEE Transactions on Intelligent Vehicles}, 8\penalty0 (6):\penalty0 3692–3711, 2023.

\bibitem[Tian et~al.(2024)Tian, Peng, Song, Jin, Yu, Mi, and Yu]{tian2024toward}
Ye Tian, Baolin Peng, Linfeng Song, Lifeng Jin, Dian Yu, Haitao Mi, and Dong Yu.
\newblock Toward self-improvement of llms via imagination, searching, and criticizing.
\newblock \emph{arXiv preprint arXiv:2404.12253}, 2024.

\bibitem[Tong et~al.(2024)Tong, Brown, Wu, Woo, Middepogu, Akula, Yang, Yang, Iyer, Pan, Wang, Fergus, LeCun, and Xie]{tong2024cambrian1fullyopenvisioncentric}
Shengbang Tong, Ellis Brown, Penghao Wu, Sanghyun Woo, Manoj Middepogu, Sai~Charitha Akula, Jihan Yang, Shusheng Yang, Adithya Iyer, Xichen Pan, Austin Wang, Rob Fergus, Yann LeCun, and Saining Xie.
\newblock Cambrian-1: A fully open, vision-centric exploration of multimodal llms, 2024.

\bibitem[Uesato et~al.(2022)Uesato, Kushman, Kumar, Song, Siegel, Wang, Creswell, Irving, and Higgins]{uesato2022solving}
Jonathan Uesato, Nate Kushman, Ramana Kumar, Francis Song, Noah Siegel, Lisa Wang, Antonia Creswell, Geoffrey Irving, and Irina Higgins.
\newblock Solving math word problems with process-and outcome-based feedback.
\newblock \emph{arXiv preprint arXiv:2211.14275}, 2022.

\bibitem[Wang et~al.(2024{\natexlab{a}})Wang, Song, Tian, Peng, Yu, Mi, Su, and Yu]{wang2024litesearch}
Ante Wang, Linfeng Song, Ye Tian, Baolin Peng, Dian Yu, Haitao Mi, Jinsong Su, and Dong Yu.
\newblock Litesearch: Efficacious tree search for llm.
\newblock \emph{arXiv preprint arXiv:2407.00320}, 2024{\natexlab{a}}.

\bibitem[Wang et~al.(2022{\natexlab{a}})Wang, Yang, Hu, Li, Lin, Gan, Liu, Liu, and Wang]{wang2022git}
Jianfeng Wang, Zhengyuan Yang, Xiaowei Hu, Linjie Li, Kevin Lin, Zhe Gan, Zicheng Liu, Ce Liu, and Lijuan Wang.
\newblock Git: A generative image-to-text transformer for vision and language.
\newblock \emph{arXiv preprint arXiv:2205.14100}, 2022{\natexlab{a}}.

\bibitem[Wang et~al.(2023{\natexlab{a}})Wang, Zhou, Xu, Shi, Zhao, Xu, Ye, Yan, Zhang, Zhu, et~al.]{wang2023evaluation}
Junyang Wang, Yiyang Zhou, Guohai Xu, Pengcheng Shi, Chenlin Zhao, Haiyang Xu, Qinghao Ye, Ming Yan, Ji Zhang, Jihua Zhu, et~al.
\newblock Evaluation and analysis of hallucination in large vision-language models.
\newblock \emph{arXiv preprint arXiv:2308.15126}, 2023{\natexlab{a}}.

\bibitem[Wang et~al.(2024{\natexlab{b}})Wang, Wang, Xu, Zhang, Gu, Jia, Wang, Xu, Yan, Zhang, and Sang]{wang2024amberllmfreemultidimensionalbenchmark}
Junyang Wang, Yuhang Wang, Guohai Xu, Jing Zhang, Yukai Gu, Haitao Jia, Jiaqi Wang, Haiyang Xu, Ming Yan, Ji Zhang, and Jitao Sang.
\newblock Amber: An llm-free multi-dimensional benchmark for mllms hallucination evaluation, 2024{\natexlab{b}}.

\bibitem[Wang et~al.(2023{\natexlab{b}})Wang, He, Li, Liu, and Lim]{wang2023mitigatingfinegrainedhallucinationfinetuning}
Lei Wang, Jiabang He, Shenshen Li, Ning Liu, and Ee-Peng Lim.
\newblock Mitigating fine-grained hallucination by fine-tuning large vision-language models with caption rewrites, 2023{\natexlab{b}}.

\bibitem[Wang et~al.(2024{\natexlab{c}})Wang, Bai, Tan, Wang, Fan, Bai, Chen, Liu, Wang, Ge, Fan, Dang, Du, Ren, Men, Liu, Zhou, Zhou, and Lin]{wang2024qwen2vlenhancingvisionlanguagemodels}
Peng Wang, Shuai Bai, Sinan Tan, Shijie Wang, Zhihao Fan, Jinze Bai, Keqin Chen, Xuejing Liu, Jialin Wang, Wenbin Ge, Yang Fan, Kai Dang, Mengfei Du, Xuancheng Ren, Rui Men, Dayiheng Liu, Chang Zhou, Jingren Zhou, and Junyang Lin.
\newblock Qwen2-vl: Enhancing vision-language model's perception of the world at any resolution, 2024{\natexlab{c}}.

\bibitem[Wang et~al.(2024{\natexlab{d}})Wang, Li, Shao, Xu, Dai, Li, Chen, Wu, and Sui]{wang2024mathshepherdverifyreinforcellms}
Peiyi Wang, Lei Li, Zhihong Shao, R.~X. Xu, Damai Dai, Yifei Li, Deli Chen, Y. Wu, and Zhifang Sui.
\newblock Math-shepherd: Verify and reinforce llms step-by-step without human annotations, 2024{\natexlab{d}}.

\bibitem[Wang et~al.(2023{\natexlab{c}})Wang, Lv, Yu, Hong, Qi, Wang, Ji, Yang, Zhao, Song, et~al.]{wang2023cogvlm}
Weihan Wang, Qingsong Lv, Wenmeng Yu, Wenyi Hong, Ji Qi, Yan Wang, Junhui Ji, Zhuoyi Yang, Lei Zhao, Xixuan Song, et~al.
\newblock Cogvlm: Visual expert for pretrained language models.
\newblock \emph{arXiv preprint arXiv:2311.03079}, 2023{\natexlab{c}}.

\bibitem[Wang et~al.(2023{\natexlab{d}})Wang, Zheng, Sun, Jia, Wongkamjan, Xu, and Huang]{wang2023coplanner}
Xiyao Wang, Ruijie Zheng, Yanchao Sun, Ruonan Jia, Wichayaporn Wongkamjan, Huazhe Xu, and Furong Huang.
\newblock Coplanner: Plan to roll out conservatively but to explore optimistically for model-based rl.
\newblock \emph{arXiv preprint arXiv:2310.07220}, 2023{\natexlab{d}}.

\bibitem[Wang et~al.(2024{\natexlab{e}})Wang, Chen, Wang, Zhou, Zhou, Yao, Zhou, Goldstein, Bhatia, Huang, et~al.]{wang2024enhancing}
Xiyao Wang, Jiuhai Chen, Zhaoyang Wang, Yuhang Zhou, Yiyang Zhou, Huaxiu Yao, Tianyi Zhou, Tom Goldstein, Parminder Bhatia, Furong Huang, et~al.
\newblock Enhancing visual-language modality alignment in large vision language models via self-improvement.
\newblock \emph{arXiv preprint arXiv:2405.15973}, 2024{\natexlab{e}}.

\bibitem[Wang et~al.(2024{\natexlab{f}})Wang, Song, Tian, Yu, Peng, Mi, Huang, and Yu]{wang2024towards}
Xiyao Wang, Linfeng Song, Ye Tian, Dian Yu, Baolin Peng, Haitao Mi, Furong Huang, and Dong Yu.
\newblock Towards self-improvement of llms via mcts: Leveraging stepwise knowledge with curriculum preference learning.
\newblock \emph{arXiv preprint arXiv:2410.06508}, 2024{\natexlab{f}}.

\bibitem[Wang et~al.(2024{\natexlab{g}})Wang, Zhou, Liu, Lu, Xu, He, Yoon, Lu, Liu, Bertasius, Bansal, Yao, and Huang]{wang-etal-2024-mementos}
Xiyao Wang, Yuhang Zhou, Xiaoyu Liu, Hongjin Lu, Yuancheng Xu, Feihong He, Jaehong Yoon, Taixi Lu, Fuxiao Liu, Gedas Bertasius, Mohit Bansal, Huaxiu Yao, and Furong Huang.
\newblock Mementos: A comprehensive benchmark for multimodal large language model reasoning over image sequences.
\newblock In \emph{Proceedings of the 62nd Annual Meeting of the Association for Computational Linguistics (Volume 1: Long Papers)}, pages 416--442. Association for Computational Linguistics, 2024{\natexlab{g}}.

\bibitem[Wang et~al.(2022{\natexlab{b}})Wang, Yu, Yu, Dai, Tsvetkov, and Cao]{wang2021simvlm}
Zirui Wang, Jiahui Yu, Adams~Wei Yu, Zihang Dai, Yulia Tsvetkov, and Yuan Cao.
\newblock Simvlm: Simple visual language model pretraining with weak supervision.
\newblock In \emph{ICLR}, 2022{\natexlab{b}}.

\bibitem[Wu et~al.(2024)Wu, Zheng, Ma, Lu, Guo, Zhang, Chen, Guo, Shen, and Zha]{wu2024lotlip}
Wei Wu, Kecheng Zheng, Shuailei Ma, Fan Lu, Yuxin Guo, Yifei Zhang, Wei Chen, Qingpei Guo, Yujun Shen, and Zheng-Jun Zha.
\newblock Lotlip: Improving language-image pre-training for long text understanding.
\newblock \emph{arXiv preprint arXiv:2410.05249}, 2024.

\bibitem[Xu et~al.(2024)Xu, Sehwag, Koppel, Zhu, An, Huang, and Ganesh]{xu2024genarm}
Yuancheng Xu, Udari~Madhushani Sehwag, Alec Koppel, Sicheng Zhu, Bang An, Furong Huang, and Sumitra Ganesh.
\newblock Genarm: Reward guided generation with autoregressive reward model for test-time alignment.
\newblock \emph{arXiv preprint arXiv:2410.08193}, 2024.

\bibitem[Xue et~al.(2024)Xue, Chen, Li, Hu, Zhu, Li, Fang, Tang, Yang, Liu, He, Yin, Molchanov, Kautz, Fan, Zhu, Lu, and Han]{xue2024longvilascalinglongcontextvisual}
Fuzhao Xue, Yukang Chen, Dacheng Li, Qinghao Hu, Ligeng Zhu, Xiuyu Li, Yunhao Fang, Haotian Tang, Shang Yang, Zhijian Liu, Ethan He, Hongxu Yin, Pavlo Molchanov, Jan Kautz, Linxi Fan, Yuke Zhu, Yao Lu, and Song Han.
\newblock Longvila: Scaling long-context visual language models for long videos, 2024.

\bibitem[Yang et~al.(2024)Yang, Zhang, Hui, Gao, Yu, Li, Liu, Tu, Zhou, Lin, Lu, Xue, Lin, Liu, Ren, and Zhang]{yang2024qwen25mathtechnicalreportmathematical}
An Yang, Beichen Zhang, Binyuan Hui, Bofei Gao, Bowen Yu, Chengpeng Li, Dayiheng Liu, Jianhong Tu, Jingren Zhou, Junyang Lin, Keming Lu, Mingfeng Xue, Runji Lin, Tianyu Liu, Xingzhang Ren, and Zhenru Zhang.
\newblock Qwen2.5-math technical report: Toward mathematical expert model via self-improvement, 2024.

\bibitem[Yang et~al.(2023)Yang, Li, Lin, Wang, Lin, Liu, and Wang]{yang2023dawn}
Zhengyuan Yang, Linjie Li, Kevin Lin, Jianfeng Wang, Chung-Ching Lin, Zicheng Liu, and Lijuan Wang.
\newblock The dawn of lmms: Preliminary explorations with gpt-4v (ision).
\newblock \emph{arXiv preprint arXiv:2309.17421}, 2023.

\bibitem[Yu et~al.(2022)Yu, Wang, Vasudevan, Yeung, Seyedhosseini, and Wu]{yu2022coca}
Jiahui Yu, Zirui Wang, Vijay Vasudevan, Legg Yeung, Mojtaba Seyedhosseini, and Yonghui Wu.
\newblock Coca: Contrastive captioners are image-text foundation models.
\newblock \emph{arXiv preprint arXiv:2205.01917}, 2022.

\bibitem[Yu et~al.(2023)Yu, Yang, Li, Wang, Lin, Liu, Wang, and Wang]{yu2023mmvetevaluatinglargemultimodal}
Weihao Yu, Zhengyuan Yang, Linjie Li, Jianfeng Wang, Kevin Lin, Zicheng Liu, Xinchao Wang, and Lijuan Wang.
\newblock Mm-vet: Evaluating large multimodal models for integrated capabilities, 2023.

\bibitem[Yuan et~al.(2021)Yuan, Chen, Chen, Codella, Dai, Gao, Hu, Huang, Li, Li, et~al.]{yuan2021florence}
Lu Yuan, Dongdong Chen, Yi-Ling Chen, Noel Codella, Xiyang Dai, Jianfeng Gao, Houdong Hu, Xuedong Huang, Boxin Li, Chunyuan Li, et~al.
\newblock Florence: A new foundation model for computer vision.
\newblock \emph{arXiv preprint arXiv:2111.11432}, 2021.

\bibitem[Yue et~al.(2024)Yue, Ni, Zhang, Zheng, Liu, Zhang, Stevens, Jiang, Ren, Sun, Wei, Yu, Yuan, Sun, Yin, Zheng, Yang, Liu, Huang, Sun, Su, and Chen]{yue2024mmmumassivemultidisciplinemultimodal}
Xiang Yue, Yuansheng Ni, Kai Zhang, Tianyu Zheng, Ruoqi Liu, Ge Zhang, Samuel Stevens, Dongfu Jiang, Weiming Ren, Yuxuan Sun, Cong Wei, Botao Yu, Ruibin Yuan, Renliang Sun, Ming Yin, Boyuan Zheng, Zhenzhu Yang, Yibo Liu, Wenhao Huang, Huan Sun, Yu Su, and Wenhu Chen.
\newblock Mmmu: A massive multi-discipline multimodal understanding and reasoning benchmark for expert agi, 2024.

\bibitem[Zhang et~al.(2024)Zhang, Zhoubian, Hu, Yue, Dong, and Tang]{zhang2024rest}
Dan Zhang, Sining Zhoubian, Ziniu Hu, Yisong Yue, Yuxiao Dong, and Jie Tang.
\newblock Rest-mcts*: Llm self-training via process reward guided tree search.
\newblock \emph{arXiv preprint arXiv:2406.03816}, 2024.

\bibitem[Zhao et~al.(2023)Zhao, Wang, Ouyang, Dong, Wang, and He]{zhao2023hallucinations}
Zhiyuan Zhao, Bin Wang, Linke Ouyang, Xiaoyi Dong, Jiaqi Wang, and Conghui He.
\newblock Beyond hallucinations: Enhancing lvlms through hallucination-aware direct preference optimization, 2023.

\bibitem[Zhou et~al.(2024{\natexlab{a}})Zhou, Cui, Rafailov, Finn, and Yao]{zhou2024aligning}
Yiyang Zhou, Chenhang Cui, Rafael Rafailov, Chelsea Finn, and Huaxiu Yao.
\newblock Aligning modalities in vision large language models via preference fine-tuning.
\newblock \emph{arXiv preprint arXiv:2402.11411}, 2024{\natexlab{a}}.

\bibitem[Zhou et~al.(2024{\natexlab{b}})Zhou, Fan, Cheng, Yang, Chen, Cui, Wang, Li, Zhang, and Yao]{zhou2024calibrated}
Yiyang Zhou, Zhiyuan Fan, Dongjie Cheng, Sihan Yang, Zhaorun Chen, Chenhang Cui, Xiyao Wang, Yun Li, Linjun Zhang, and Huaxiu Yao.
\newblock Calibrated self-rewarding vision language models.
\newblock \emph{arXiv preprint arXiv:2405.14622}, 2024{\natexlab{b}}.

\end{thebibliography}
}

\clearpage
\appendix
\setcounter{page}{1}

\twocolumn[{\renewcommand\twocolumn[1][]{#1}
\maketitlesupplementary

\begin{tcolorbox} 
\centering
\small
\begin{tabular}{p{0.96\textwidth}}
 {\bf GPT-4o and human evaluation prompt} \vspace{1mm} \\
Evaluate the following image captions generated by two vision-language models (VLMs) in response to a given image. Please decide which caption is ``better" based on the following criteria:\vspace{2mm} \\
Richness of Content: The caption should provide a comprehensive description of the visual elements in the image, detailing objects, actions, colors, or settings that enhance the viewer's understanding of the scene. \\
Accuracy: The caption must accurately reflect what is in the image without adding incorrect information or making erroneous assumptions about unseen aspects. \\
Harmlessness and Appropriateness: The caption should be free from any potentially harmful, offensive, or inappropriate content. Ensure that the caption does not make unnecessary assumptions about people’s identities, backgrounds, or intentions. \\
Creativity and Elaboration: If the caption can remain accurate and harmless, it should include suitable and imaginative elaborations that enrich the scene's description (e.g., suggesting possible actions, moods, or a gentle interpretation of the scene’s context). \\
Clarity and Coherence: The caption should be clear, concise, and well-structured, helping any viewer quickly understand the image’s content without ambiguity. \vspace{2mm} \\
After considering these criteria, please directly output your choice, the format of your choice should be `Response1 is better', `Response2 is better' or `Tie'. \\
Response1: \{\} \\
Response2: \{\} \\
\end{tabular}
\end{tcolorbox}
\label{tab:example_pairwise_prompts}
\captionof{table}{Prompt used for image caption quality evaluation with GPT-4o and human.}
\vspace{1em}
}]

\section{Evaluation prompt for GPT and human evaluation}
\label{app: A}

In this section, we provide the detailed prompt for GPT-4o and human evaluation in Section~\ref{sec_41}. 
We evaluate the caption quality from 5 aspects: Richness of Content, Accuracy, Harmlessness and Appropriateness, Creativity and Elaboration, Clarity and Coherence.

\section{Details of \ours training}
\label{app: C}

Our training dataset consists of 378k $<$current sentence, current sentence clip score, next sentence, image$>$. Based on this training set, we train \ours for 3 epochs with a learning rate of 5e-5 and a batch size of 1024. The latent dimension of the value head in \ours is 2560. $\gamma$ used for TD learning is 0.9. The entire training process is conducted on 8$\times$80G A100 GPUs for 50 hours.

\section{More experiments}
\label{app: more_exp}

In this section, we provide comparison with additional various finetuning (HADPO~\cite{zhao2023hallucinations}, POVID~\cite{zhou2024aligning}, CSR~\cite{zhou2024calibrated}) and guided decoding (VCD~\cite{leng2023mitigatingobjecthallucinationslarge}, CLIP-Guided) methods. Since all previous methods use LLaVA-1.5-7B as the base model, we also experiment on LLaVA-1.5-7B for generating description data and training VisVM with CLIP-PRM. The experimental results in Table~\ref{table: app-more-baseline} show that \ours consistently outperforms all other methods across all metrics. Furthermore, \ours focuses on inference-time search, making the improvement orthogonal to model finetuning, and capable of further enhancing model performance during inference.

\begin{table*}[t]
\centering
\caption{Comparison of \ours-guided search with various finetuning and decoding methods. \ours significantly outperforms all other methods across all metrics.}
\resizebox{\linewidth}{!}{%
\begin{tabular}{l|l|ccccc}
\toprule 
Base & Finetuning/Searching Method & CHAIRs $\downarrow$ & CHAIRi $\downarrow$ & MMHal $\uparrow$ & AMBER Cov $\uparrow$\\
\midrule
\multirow{7}{*}{LLaVA-1.5-7B}   & Greedy (Default)    & 46.2 & 10.1 & 2.13 & 49.7 \\
                                & HADPO  (Finetuning) & 43.6 & 9.7  & 2.27 & 48.9\\
                                & POVID  (Finetuning) & 39.7 & 9.4  & 2.45 & 46.8\\
                                & CSR    (Finetuning) & 35.2 & 8.6  & 2.69 & 44.3 \\
                                & VCD    (Searching)  & 33.4 & 8.9  & 2.45 & 51.9 \\
                                & CLIP-Guided (Searching)         & 32.2 & 8.4  & 2.72 & 52.4 \\
                                \rowcolor{gray!30}
\cellcolor{white}               & \textbf{\ours-Guided}       & 30.4 & 8.2  & 2.75 & 56.1 \\
\bottomrule
\end{tabular}
}
\label{table: app-more-baseline}
\end{table*} 

\section{A case study for \ours Analysis}
\label{app: Analysis}

\begin{figure*}[!t]
    \centering
    \includegraphics[width=0.9\linewidth]{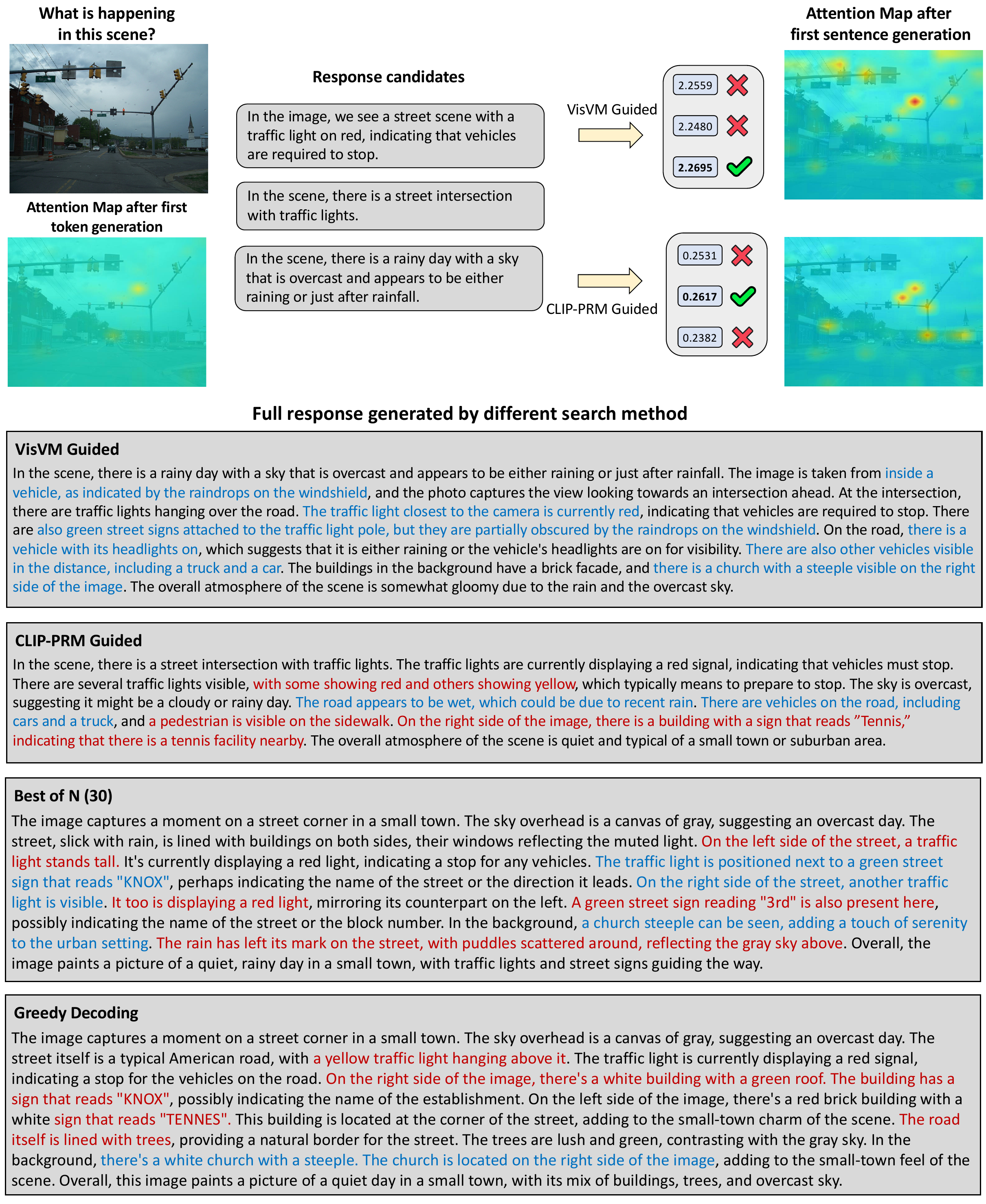} 
    \caption{A case study on \ours-guided search. The upper part shows how \ours and CLIP-PRM make different choices when given the same step response candidates, and the changes in the LLaVA-Next attention map after the choices are made. We can observe significant differences in the attention maps. The second part presents the complete responses obtained using different search methods, with \textcolor[HTML]{0070C0}{blue text} indicating correct details and \textcolor[HTML]{C00000}{red text} indicating hallucinations. \ours-guided search obtains response with richer details and fewer hallucinations.}
    \label{fig:explain_fig}
    \vspace{-5pt}
\end{figure*}

To better understand how \ours influences VLM's response generation, this section examines how \ours and CLIP-PRM select responses when presented with the same set of candidates. As illustrated in the upper part of Figure~\ref{fig:explain_fig}, given an image and prompt, LLAVA-Next generated three different response candidates. When using CLIP to directly score the responses, the second candidate received the highest clip score which is 0.2617 and is thus chosen as the current step response. However, \ours considers potential hallucinations in subsequent responses induced by the sentence, resulting in a higher value 2.2695 for the third sentence among the given candidates. Therefore, the third sentence is selected. We search and obtain the final complete response using both CLIP-PRM and \ours following their choices, as shown in the lower part of Figure~\ref{fig:explain_fig}. We observe that the response generated through \ours search indeed contains more details and less hallucinations.

Additionally, Figure~\ref{fig:explain_fig} also presents the changes in the LLAVA-Next's image-text cross-attention map following the selection of different candidates. We note significant differences in the attention maps after choosing different candidates. \ours's attention map more comprehensively covers the entire image, enabling it to catch visual details such as ``water droplets obscuring the road sign.'' In contrast, the CLIP-PRM's attention map over-emphasizes the area around the traffic light, leading to inaccurate description of the light's color and missing other visual details.

\section{More case studies}
\label{app: B}
In this section, we give more case studies from Table~\ref{apd: case1} to Table~\ref{apd: case9} to compare \ours guided decoding results and CLIP-PRM guided decoding results.

\begin{table*}[!htbp]
\begin{minipage}{0.98\linewidth}
\centering
\scalebox{0.98}{
\begin{tabular}{c p{0.83\linewidth}}
\toprule
 \multicolumn{2}{l}{\bf Example 1}  \\
\midrule
 & \includegraphics[height=5cm]{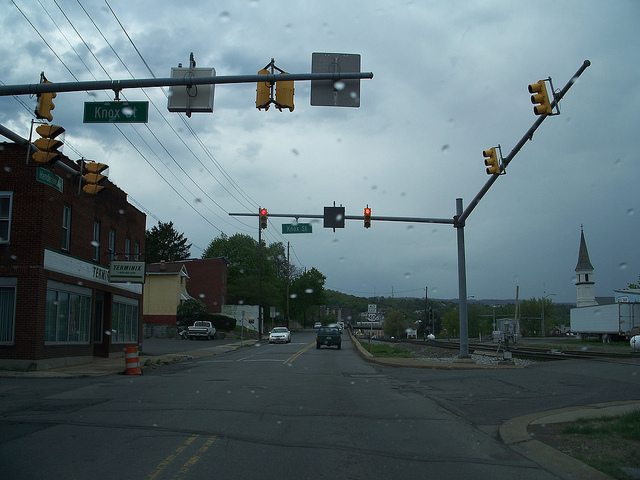} \\

Question & What is happening in this scene? \\
\midrule
\ours Guided & In the scene, there is a rainy day with a sky that is overcast and appears to be either raining or just after rainfall. \textcolor[HTML]{0070C0}{The image is taken from inside a vehicle, as indicated by the raindrops on the windshield}, and the photo captures the view looking towards an intersection ahead.

At the intersection, there are traffic lights hanging over the road. The traffic light closest to the camera is currently red, indicating that vehicles are required to stop. \textcolor[HTML]{0070C0}{There are also green street signs attached to the traffic light pole, but they are partially obscured by the raindrops on the windshield}.

\textcolor[HTML]{0070C0}{On the road, there is a vehicle with its headlights on}, which suggests that it is either raining or the vehicle's headlights are on for visibility. There are also other vehicles visible in the distance, including a truck and a car.

The buildings in the background have a brick facade, and \textcolor[HTML]{0070C0}{there is a church with a steeple visible on the right side of the image}. The overall atmosphere of the scene is somewhat gloomy due to the rain and the overcast sky. \\
\midrule
CLIP-PRM Guided & In the scene, there is a street intersection with traffic lights. The traffic lights are currently displaying a red signal, indicating that vehicles must stop. There are several traffic lights visible, with some showing red and others showing yellow, which typically means to prepare to stop.

The sky is overcast, suggesting it might be a cloudy or rainy day. The road appears to be wet, which could be due to recent rain. There are vehicles on the road, including cars and a truck, and a pedestrian is visible on the sidewalk.

On the right side of the image, there is a building with a sign that reads "Tennis," indicating that there is a tennis facility nearby. The overall atmosphere of the scene is quiet and typical of a small town or suburban area.   \\
\bottomrule
\end{tabular}
}
\vspace{0mm}
\captionof{table}{}
\label{apd: case1}  
\end{minipage}
\end{table*}

\begin{table*}[!htbp]
\begin{minipage}{0.98\linewidth}
\centering
\scalebox{0.98}{
\begin{tabular}{c p{0.83\linewidth}}
\toprule
 \multicolumn{2}{l}{\bf Example 2}  \\
\midrule
 & \includegraphics[height=5cm]{figures/000000003817.jpg} \\

Question & What is happening in this scene? \\
\midrule
\ours Guided & In the image, there is a person riding a motorcycle or moped with a large bundle of bananas tied to the back. The person is wearing a helmet and appears to be on a rural road, with lush greenery and trees on both sides, suggesting a tropical or subtropical region. \textcolor[HTML]{0070C0}{There are other people in the background, some of whom are walking along the road, and there's a small building or structure visible in the distance.} \textcolor[HTML]{0070C0}{The road seems to be a secondary road, with yellow markings indicating it's a two-way road}, and \textcolor[HTML]{0070C0}{there's a sign or symbol on the right side of the road that is partially obscured}. The overall scene suggests a day-to-day activity in a rural or semi-rural area, possibly related to transportation of goods or agricultural produce.  \\
\midrule
CLIP-PRM Guided & In the image, there is a person riding a motorcycle loaded with a large quantity of green bananas. The bananas are tied to the back of the motorcycle, and the rider appears to be transporting them down a road. The setting is rural, with trees and greenery surrounding the road, suggesting that the bananas might have been harvested from nearby plantations. The person is wearing a helmet, which is a good safety practice when riding a motorcycle. The scene captures a moment of daily life in a region where bananas are a significant crop.   \\
\bottomrule
\end{tabular}
}
\vspace{0mm}
\captionof{table}{}
\label{apd: case2}   
\end{minipage}
\end{table*}

\begin{table*}[!htbp]
\begin{minipage}{0.98\linewidth}
\centering
\scalebox{0.98}{
\begin{tabular}{c p{0.83\linewidth}}
\toprule
 \multicolumn{2}{l}{\bf Example 3}  \\
\midrule
 & \includegraphics[height=5cm]{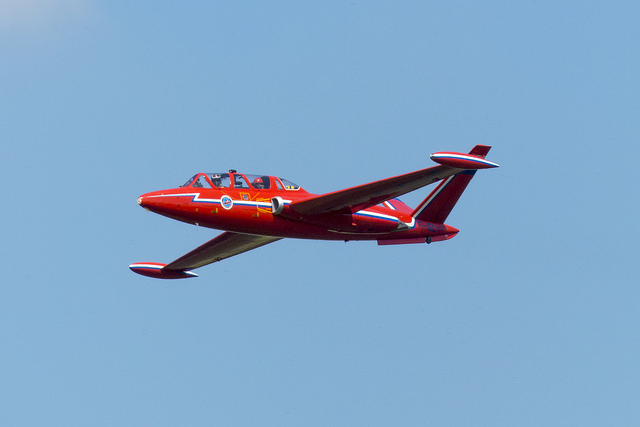} \\

Question & Can you describe the main features of this image for me? \\
\midrule
\ours Guided & The image shows a single-engine, propeller-driven aircraft flying against a clear blue sky. The aircraft has a bright red fuselage with \textcolor[HTML]{0070C0}{white and blue stripes running along its side}, and \textcolor[HTML]{0070C0}{a white and blue striped tail}. There are no visible registration numbers or other identifying marks on the aircraft. \textcolor[HTML]{0070C0}{The aircraft appears to be a small personal or recreational plane}, given its size and the absence of any visible branding or commercial markings. The wings are swept back slightly, and there are no visible landing gear, suggesting that the aircraft is either in flight or has just taken off. The image captures the aircraft from a perspective below and to the side, providing a clear view of its underside and tail section. \\
\midrule
CLIP-PRM Guided & The image shows a red airplane flying against a clear blue sky. The aircraft has a single propeller and a tail fin with a white stripe. It appears to be a small, single-engine plane, possibly a vintage or classic model, given its design and color scheme. The plane is captured in mid-flight, with its nose pointed slightly upward. The wings are swept back, which is a common feature in many small aircraft for improved aerodynamics. There are no visible windows on the fuselage, suggesting it may be a one-seater or a light utility aircraft. The overall impression is that of a leisurely flight or a demonstration of the aircraft's capabilities.  \\
\bottomrule
\end{tabular}
}
\vspace{0mm}
\captionof{table}{}
\label{apd: case3}   
\end{minipage}
\end{table*}

\begin{table*}[!htbp]
\begin{minipage}{0.98\linewidth}
\centering
\scalebox{0.98}{
\begin{tabular}{c p{0.83\linewidth}}
\toprule
 \multicolumn{2}{l}{\bf Example 4}  \\
\midrule
 & \includegraphics[height=5cm]{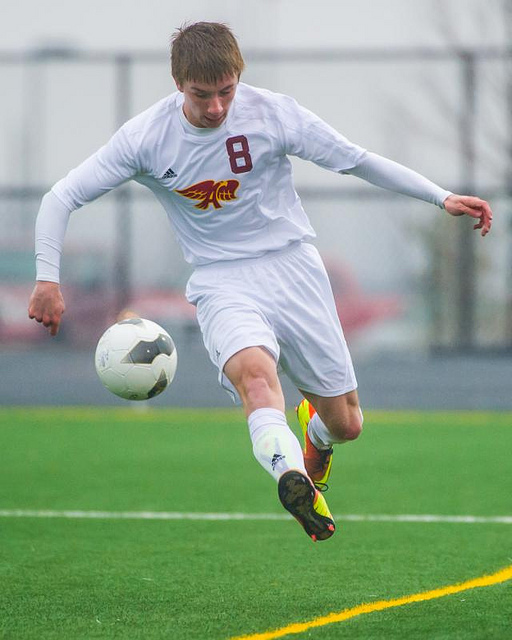} \\

Question & Can you elaborate on the elements of the picture provided? \\
\midrule
\ours Guided & In the image, there is a soccer player in action on a field. The player is wearing a white soccer kit with a red number "8" on the front, which is a common way to identify a player's position or role within a team. \textcolor[HTML]{0070C0}{The jersey also has an emblem that appears to be a bird, possibly a falcon, which could be the team's logo}. The player is wearing white socks with black cleats, which are typical for soccer players. The player is in the process of kicking a soccer ball, which is white with black pentagons, known as a pentagon ball, which is the standard shape for soccer balls. The player\'s body is in a dynamic pose, with one leg extended forward and the other bent at the knee, indicating a powerful kick. \textcolor[HTML]{0070C0}{The field is marked with white lines, which are standard for soccer fields, and there is a yellow line on the side, which could be the touchline or the goal line}, depending on the specific context of the game. In the background, \textcolor[HTML]{0070C0}{there is a chain-link fence, which is common for soccer fields to keep the ball from going out of play}. \textcolor[HTML]{0070C0}{There is also a vehicle, possibly a car or van, partially visible behind the fence}, suggesting that the field might be located in a park or recreational area near a road. The sky is overcast, and the lighting suggests it might be late afternoon or early evening, as the shadows are long and the light is diffused. The overall atmosphere of the image is one of action and competition, capturing a moment of athletic endeavor. \\
\midrule
CLIP-PRM Guided & In the image, we see a soccer player in action on a field. The player is wearing a white uniform with a red number "8" on the front, which is likely his team\'s jersey number. The uniform is the standard soccer attire, with shorts, a shirt, and possibly socks and shin guards that are not fully visible. The player is wearing yellow and orange cleats, which are common for soccer players to provide traction on the field. The player is in the process of kicking the soccer ball, which is in mid-air, indicating a dynamic moment in the game. The player\'s right foot is extended towards the ball, suggesting a powerful strike. The player's body is positioned with one leg bent and the other extended, which is a common stance for a player about to kick the ball. The background of the image shows a soccer field with a goal post visible, indicating that the player is near the goal area. The field appears to be well-maintained with clear boundary lines, and there are no visible spectators or other players, which puts the focus on the player and his action. Overall, the image captures a snapshot of an intense moment during a soccer game. The athlete's concentration and the positioning of the ball and the player\'s leg suggest a high level of skill and athleticism.  \\
\bottomrule
\end{tabular}
}
\vspace{0mm}
\captionof{table}{}
\label{apd: case4}     
\end{minipage}
\end{table*}

\begin{table*}[!htbp]
\begin{minipage}{0.98\linewidth}
\centering
\scalebox{0.98}{
\begin{tabular}{c p{0.83\linewidth}}
\toprule
 \multicolumn{2}{l}{\bf Example 5}  \\
\midrule
 & \includegraphics[height=5cm]{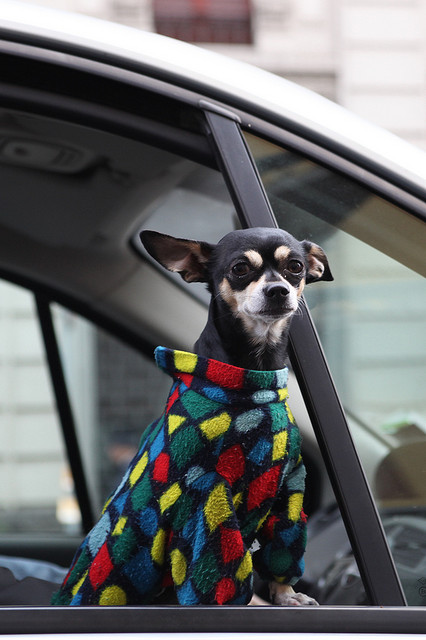} \\

Question & Write a detailed description of the given image. \\
\midrule
\ours Guided & In the heart of a bustling city, a small Chihuahua with a vibrant coat of black and brown fur is the star of this scene. The dog, standing on its hind legs, has its front paws resting on the window sill of a white car. The car is parked on the side of a street, with a building visible in the background. The dog is dressed in \textcolor[HTML]{0070C0}{a sweater that is a riot of colors - red, blue, green, and yellow.} \textcolor[HTML]{0070C0}{The sweater, made of soft fabric}, adds a touch of whimsy to the otherwise mundane urban setting.\ The dog's head is tilted to the side, as if it's curiously observing its surroundings. \textcolor[HTML]{0070C0}{Its eyes are wide open}, taking in the sights of the city. Despite being inside a car, the dog seems to be enjoying the view outside. The image captures a moment of joy and curiosity in an everyday urban scene. It's a reminder that even in the hustle and bustle of city life, there's always room for a little bit of fun and playfulness.  \\
\midrule
CLIP-PRM Guided & In the heart of a bustling city, a small Chihuahua has found its perch on the passenger side of a car. The dog, dressed in a vibrant sweater, is the star of this urban tableau. The sweater, a riot of colors, is adorned with a playful polka dot pattern that stands out against the dog's dark fur. The dog is seated comfortably on the car's door sill, its front paws resting on the window ledge. Its head is tilted slightly to the left, as if it's curiously observing the world outside. The window itself is rolled down, allowing the dog to enjoy the sights and sounds of the city. The car itself is a stark white, contrasting with the dog's colorful attire. It's parked on a street lined with buildings, their details blurred by the distance. The overall scene paints a picture of a day in the life of this charming Chihuahua.  \\
\bottomrule
\end{tabular}
}
\vspace{0mm}
\captionof{table}{}
\label{apd: case5}    
\end{minipage}
\end{table*}

\begin{table*}[!htbp]
\begin{minipage}{0.98\linewidth}
\centering
\scalebox{0.98}{
\begin{tabular}{c p{0.83\linewidth}}
\toprule
 \multicolumn{2}{l}{\bf Example 6}  \\
\midrule
 & \includegraphics[height=5cm]{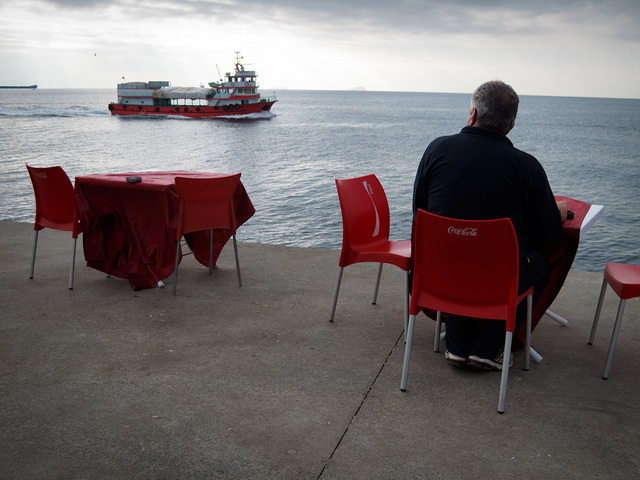} \\

Question & Can you describe the main features of this image for me? \\
\midrule
\ours Guided & The image depicts a serene outdoor setting with a view of the ocean. In the foreground, there is a man seated at a table with a red chair. He appears to be looking out at the water, possibly enjoying the view or perhaps deep in thought. The table is covered with a red cloth, and \textcolor[HTML]{0070C0}{there is a Coca-Cola logo on the chair}, suggesting the location might be a restaurant or café with branded furniture. In the background, there is a large red and white boat on the water, which \textcolor[HTML]{0070C0}{seems to be in motion, as suggested by the wake it is leaving behind}. The sky is overcast, with a soft light that suggests either dawn or dusk. The water is calm, reflecting the colors of the sky and the boat. The overall scene is one of tranquility and solitude, with the man at the table as the main subject, framed by the expansive view of the ocean and the boat. The red color of the tablecloth and chair adds a vibrant contrast to the cooler tones of the water and sky.  \\
\midrule
CLIP-PRM Guided & The image shows a person sitting at a table on what appears to be a pier or waterfront. The person is facing away from the camera, looking out towards the water. The table is set with a red tablecloth and there are two red chairs, one of which the person is occupying. In the background, there is a large body of water, and on the water, there is a red and white boat that seems to be moving away from the pier. The sky is overcast, suggesting either early morning or late afternoon light conditions. The overall atmosphere of the image is calm and serene, with a sense of solitude and contemplation.  \\
\bottomrule
\end{tabular}
}
\vspace{0mm}
\captionof{table}{}
\label{apd: case6}    
\end{minipage}
\end{table*}

\begin{table*}[!htbp]
\begin{minipage}{0.98\linewidth}
\centering
\scalebox{0.98}{
\begin{tabular}{c p{0.83\linewidth}}
\toprule
 \multicolumn{2}{l}{\bf Example 7}  \\
\midrule
 & \includegraphics[height=5cm]{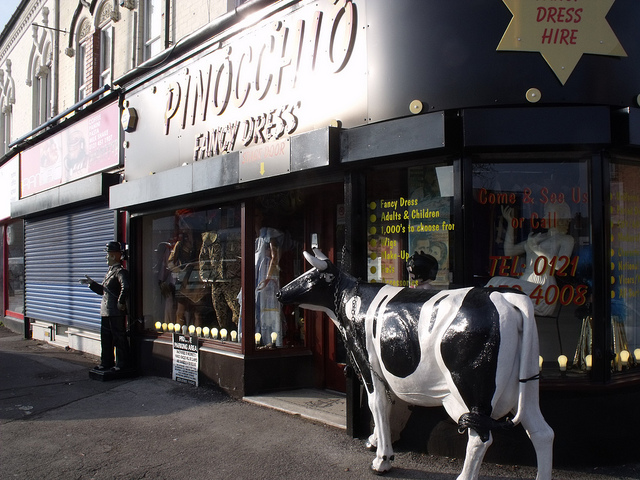} \\

Question & What's happening in the scene? \\
\midrule
\ours Guided & In the scene, there's a store named \textcolor[HTML]{0070C0}{"Pinocchio Fancy Dress"} which suggests it sells costumes or special occasion outfits. \textcolor[HTML]{0070C0}{There's a display in the window with various mannequins dressed in different styles of clothing}, which might be used to showcase the types of outfits available inside the store. In front of the store, there's a statue of a large cow with a spotted pattern, possibly representing a character from a story or just a decorative element. The cow is positioned in such a way that it appears to be walking towards the store, which adds a whimsical or playful touch to the storefront. On the right side of the image, there's a sign that reads "Dress Hire" indicating that the store offers rental services for costumes or dresses. \textcolor[HTML]{0070C0}{There are also some signs with phone numbers and possibly some text that is too small to read clearly, likely providing additional information about the store's offerings or contact details.} The overall impression is that this is a storefront with a creative and engaging display designed to attract customers and convey the type of business it operates.  \\
\midrule
CLIP-PRM Guided & In the scene, there is a life-size statue of a cow with a black and white color pattern, standing in front of a store named "Pino's Fancy Dress." The storefront has a sign that reads "Dress Hire" and there are other signs indicating the services offered, such as "Fancy Dress," "Adult \& Children's Fancy Dress," and "100's Fancy Dress." The cow statue appears to be a decorative element to attract attention to the store. To the left of the cow statue, there is a mannequin dressed in what appears to be a traditional or historical costume, which could be related to the dress hire services offered by the store. The overall setting suggests this is a shop that rents costumes for various occasions, and the cow statue could be a fun and quirky way to draw customers in.  \\
\bottomrule
\end{tabular}
}
\vspace{0mm}
\captionof{table}{}
\label{apd: case7}    
\end{minipage}
\end{table*}

\begin{table*}[!htbp]
\begin{minipage}{0.98\linewidth}
\centering
\scalebox{0.98}{
\begin{tabular}{c p{0.83\linewidth}}
\toprule
 \multicolumn{2}{l}{\bf Example 8}  \\
\midrule
 & \includegraphics[height=5cm]{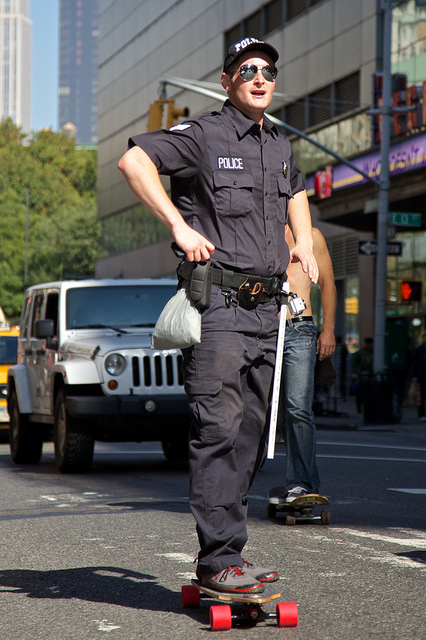} \\

Question & Can you elaborate on the elements of the picture provided? \\
\midrule
\ours Guided & In the image, there are several elements that contribute to the urban scene. The most prominent figure is a police officer standing in the middle of the street. The officer is wearing a dark-colored uniform with the word "POLICE" printed on the chest, indicating their role. They are also wearing \textcolor[HTML]{0070C0}{a cap, sunglasses, and a holster}, which are standard items of police attire. The officer is standing on a skateboard, which is a common mode of transportation in urban areas. This action could suggest that the officer is taking a break from duty or perhaps patrolling an area where skateboarding is allowed or encouraged. In the background, there is a busy street with vehicles, \textcolor[HTML]{0070C0}{including a white SUV and a yellow taxi cab}, which are typical of city traffic. There are also other pedestrians visible, contributing to the bustling atmosphere of the city. The architecture of the buildings suggests a modern cityscape, with high-rise buildings that are common in many urban centers. The presence of traffic lights and street signs further emphasizes the urban setting. The image captures a moment of everyday life in a city, with the police officer on a skateboard adding a unique and dynamic element to the scene.  \\
\midrule
CLIP-PRM Guided & The image shows a police officer riding a skateboard in a city street. The officer is wearing a uniform that includes a dark-colored shirt with the word "POLICE" on the front, and he\'s also wearing a cap with a badge. He has a walkie-talkie on his belt and is holding a bag, possibly containing personal items or equipment.\ The officer is on a skateboard, which is a common mode of transportation for urban areas. The skateboard is red and black, and the officer is wearing casual shoes and sunglasses. In the background, there are other people and vehicles, indicating that this is a busy urban environment. There's a white SUV to the left of the officer, and a traffic light is visible in the distance. The architecture suggests a modern cityscape, and the presence of a building with a sign that reads "BROOKLYN" indicates that this scene might be taking place in a city that has a borough named Brooklyn. Overall, the image captures a moment of urban law enforcement using a skateboard as a means of navigating through the city. It\'s a snapshot of a unique approach to policing in a contemporary urban setting.  \\
\bottomrule
\end{tabular}
}
\vspace{0mm}
\captionof{table}{}
\label{apd: case8}    
\end{minipage}
\end{table*}

\begin{table*}[!htbp]
\begin{minipage}{0.98\linewidth}
\centering
\scalebox{0.98}{
\begin{tabular}{c p{0.83\linewidth}}
\toprule
 \multicolumn{2}{l}{\bf Example 9}  \\
\midrule
 & \includegraphics[height=5cm]{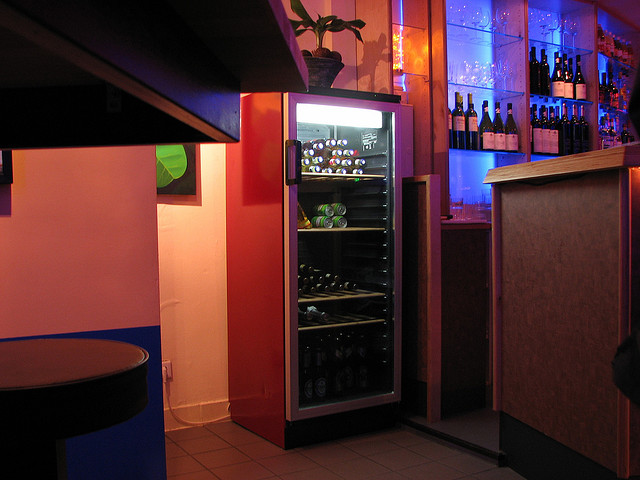} \\

Question & Can you elaborate on the elements of the picture provided? \\
\midrule
\ours Guided & In this snapshot, it appears to be a scene from a bar or a restaurant with a focus on the beverage selection. The refrigerator, which is likely a beer cooler, is prominently stocked with various bottles of what seems to be beer. The lighting is dim, with a warm glow coming from the refrigerator, which is typical for establishments that want to highlight their beverage offerings. \textcolor[HTML]{0070C0}{The presence of a potted plant and a shelf with more bottles in the background} suggests a well-decorated and inviting space. It's possible that the image was taken during a quiet moment or just before or after opening or closing hours when the establishment is not busy.  \\
\midrule
CLIP-PRM Guided & In this snapshot, it appears to be an interior view of a bar or a restaurant with a focus on a refrigerator. The refrigerator is stocked with various bottles, likely containing beverages such as beer, wine, or cocktails. The lighting in the room is dim, with the refrigerator illuminated, which suggests that it might be evening or the place is intentionally dimly lit. The presence of wine glasses on a shelf above the refrigerator indicates that the establishment serves alcoholic beverages. The ambiance seems to be casual and relaxed, possibly catering to a social or dining experience.  \\
\bottomrule
\end{tabular}
}
\vspace{0mm}
\captionof{table}{}
\label{apd: case9}   
\end{minipage}
\end{table*}

\end{document}